\documentclass[fleqn,11pt]{wlscirep}

\usepackage[utf8]{inputenc} 
\usepackage[T1]{fontenc}    
\usepackage{hyperref}       
\usepackage{url}            
\usepackage{booktabs}       
\usepackage{amsfonts}       
\usepackage{nicefrac}       
\usepackage{microtype}      
\usepackage{xcolor}         
\usepackage{helvet}         
\usepackage{graphicx}
\usepackage{placeins}
\usepackage{tikz}
\usetikzlibrary{arrows.meta, positioning}
\usepackage{soul}           
\newcommand{\fillme}[1]{#1}

\newif\ifshowrevisions
\showrevisionsfalse
\newcommand{\revised}[1]{\ifshowrevisions\textcolor{revisionred}{#1}\else#1\fi}
\newcommand{\cut}[1]{\ifshowrevisions\textcolor{revisionred}{\st{#1}}\else\fi}
\definecolor{revisionred}{HTML}{000000}

\DeclareMathAlphabet{\mathcal}{OMS}{cmsy}{m}{n}
\SetMathAlphabet{\mathcal}{bold}{OMS}{cmsy}{b}{n}
\DeclareMathOperator*{\argmax}{arg\,max}

\title{Learning aligned EEG representations with subject-specific encoders}

\author[1,2,6]{Bruna J. Lopes}
\author[1]{Gabriel Schwartz}
\author[2]{Sylvain Chevallier}
\author[1,3]{Raphael Y. de Camargo}
\author[2,3,4,5]{Bruno Aristimunha}

\affil[1]{University of S\~ao Paulo, S\~ao Paulo, Brazil}
\affil[2]{Universit\'e Paris-Saclay, Inria TAU team, LISN-CNRS, France}
\affil[3]{Federal University of ABC (UFABC), Santo Andr\'e, Brazil}
\affil[4]{Yneuro, Paris, France}
\affil[5]{Swartz Center for Computational Neuroscience (SCCN), Institute for Neural Computation (INC), University of California San Diego, La Jolla, USA}
\affil[6]{Institut de neuromodulation, GHU Paris, psychiatrie et neurosciences, centre hospitalier Sainte-Anne, p\^ole hospitalo-universitaire 15, Universit\'e Paris Cit\'e, Paris, France\\
Corresponding author: brunajaflopes@gmail.com}

\keywords{EEG decoding; brain--computer interface; cross-subject transfer; domain adaptation; deep learning; motor imagery}

\begin{abstract}

Cross-subject EEG decoding promises more training data, but it also exposes neural networks to strong inter-subject distribution shifts.
We study whether task supervision and architecture alone can learn subject-aligned representations.
We replace a shared EEG encoder with subject-specific encoders followed by a common classifier, and compare this hybrid model with standard EEGNet, AttentionBaseNet, and CTNet baselines with Euclidean Alignment (EA) on three motor-imagery datasets and one motor-execution dataset. 
EA improves shared encoders by recentering subject covariances, whereas the hybrid encoder reduces reliance on EA: removing EA has little effect on validation-loss dynamics or latent-space organization, and both hybrid variants consistently outperform non-aligned shared baselines.
Subject-specific heads increase class distinctiveness and place each subject close to its own latent manifold while improving within-subject class separation. However, on cross-subject classification, subject-specific heads hinder direct parameter transfer to unseen subjects, motivating quantitative head selection and a brief calibration session. Although decoding gains depend on the dataset and backbone, our main findings concern the solely use of architecture pressure promotes representation learning and alignment in a direction complementary to domain adaptation methods such as Euclidean Alignment.
\revised{A per-subject low-rank adapter of only $2\,C\,r$ parameters recovers the full encoder's accuracy across five backbones and ranks $r=1$ to $16$, so the per-subject module can be compressed by two to three orders of magnitude.}

\end{abstract}
\begin{document}


\flushbottom
\maketitle
\thispagestyle{empty}

\section{Introduction}

Brain decoding is an exciting yet challenging application of machine learning that aims at translating recordings of brain activity into their corresponding mental states~\cite{Xu2021BrainEncodingDecodingEEG, guetschel2023transfer}.
These systems have broad applications, including emotion recognition, sleep staging, and brain-computer interfaces (BCIs). In the latter, the objective is often to decode imagined movements from brain signals, typically represented using electroencephalography (EEG) because of its portability, high temporal resolution, and relatively low cost, making it well-suited for practical use~\cite{Roy2019}.

Traditional EEG decoding pipelines divide this process into preprocessing, feature extraction, and classification stages, relying heavily on task-specific engineering decisions rather than data-driven representation extraction~\cite{MDRM, Ang2012FBCSP, lotte:hal-01167515}.
Separating feature extraction from the main classification objective may discard task-relevant information before the predictive objective can shape the representation, and could benefit from a more automatic approach.
Deep learning techniques are currently driving a shift in this direction~\cite{ludwig2024eegminer}.
Initially established in computer vision, deep learning is an automatic feature extractor that learns discriminative representations directly from raw data~\cite{Banville2019RL}.
Applied to EEG, end-to-end architectures jointly optimize representation learning and classification, producing compact, task-relevant embeddings without handcrafted features.

Despite this progress, deep learning performance scales with the volume of available training data \cite{banville2025scalinglawsdecodingimages}.
This collides with a primary bottleneck in EEG-based brain decoding: data scarcity~\cite{aristimunha2023evaluating}.
Because each subject contributes few EEG trials~\cite{Han2023EEGGraphTransfer}, reliable models usually need data from multiple subjects or datasets.

Aggregating EEG data, however, introduces new challenges.
Different datasets frequently use distinct recording hardware, montages, and sampling frequencies, making standardization difficult~\cite{Mellot2024PhysicsinformedAU}.
EEG activity patterns themselves depend on the neuroanatomy and cognitive state of the individual at the instant of the recording, making them highly subject-specific~\cite{guetschel2023transfer, khazem2021minimizing, marty2026selfsupervised}.
Because deep neural networks typically assume that all training and test samples are drawn from the same underlying distribution, models trained using multiple subjects often suffer from \emph{negative transfer}~\cite{Pan2010, wang2019negativetransfer}, failing to learn a unified mapping because the training data lie on different manifolds. The field therefore faces a trade-off: multi-subject data is needed for scale, but pooling subjects can damage transfer. 

Domain adaptation methods aim to reduce these shifts, with some acting at the preprocessing level~\cite{He2020:euclidean, mellot:hal-04328670, bleuze2022tangent}.
For instance, Riemannian geometry-based techniques~\cite{rodrigues18:procrutes, mellot:hal-04328670, bleuze2022tangent} apply linear and non-linear transformations on the covariance representation of samples, but operate in covariance space and do not naturally extend to raw time-series inputs.
Euclidean Alignment~\cite{He2020:euclidean} is a technique designed for EEG time series, but it still consists of a simple rigid recentering of the data~\cite{junqueira2024ea}, which, while effective, may be insufficient to capture the full complexity of inter-subject variability. \revised{Complementary lines of work scale pretraining rather than trying to explicitly mitigate domain shifts, building large end-to-end EEG foundation models that learn generic representations from heterogeneous corpora and transfer them downstream~\cite{jiang2024labram, wang2024eegpt, aziz2025bcinetv1, tokenization2025hybrid}}

End-to-end training also allows deep domain adaptation (DDA) methods to place alignment inside the learning objective~\cite{Wei2021, Phunruangsakao2022DDA, Zhou2023DeepDomainEEG, Zhong2023DeepCORALEEG, bakas2025latent, klein2025mitigating}.
Zhong et al.~\cite{Zhong2023DeepCORALEEG} introduce a correlation alignment (CORAL)-based loss for minimizing statistical differences, while other methods use an adversarial objective to make representations domain-invariant~\cite{Zhou2023DeepDomainEEG, Phunruangsakao2022DDA}.
These methods, however, assume that target data is available during training, which is not always feasible to obtain. The work of Wei et al.~\cite{Wei2021} also explores Deep domain adaptation, developing a Separate-Common-Separate (SCSN) modification to Shallow FBCSP Net\cite{Schirrmeister2017} to extract common features across domain subjects. This work, however, focuses solely on training adaptation, not extending its analysis to a novel target distribution. Closest to our motivation, \cite{bakas2025latent} align subjects in the model's latent space rather than at the input; we instead factorise this latent alignment into per-subject encoders trained end-to-end.

Building on these ideas, we ask whether task supervision alone can induce cross-subject alignment when the architecture explicitly factorizes subject-specific transformations.
Inspired by the standard EEG processing pipeline, we decompose the common \emph{feature extractor–classifier} deep learning model structure by introducing subject-specific encoder heads that project each individual's EEG into a shared latent space, which is then passed to a common classification module.
The aim is to learn a subject-tailored filter for each training subject. 

This structure imposes a strong inductive bias: each head projects the subject into an individual manifold, while the joint classification objective creates pressure to align those projections for a unified classification space.
Beyond cross-subject decoding performance, we aim to study how this architectural and task pressure reshapes the geometry of the latent space and the conditional class distributions, and we examine how such a structure can be adapted at test time for unseen subjects, along with the limitations of the design.
We find that EA yields per-subject accuracy gains across backbones, \revised{and that subject-specific encoders are able to learn well class-aligned separate representations, while preserving some subject identity} from task supervision alone, and that they sharpen class distinctiveness for source subjects while leaving head selection for unseen subjects as the main bottleneck. \revised{We further show that this accuracy does not require a full per-subject encoder: a low-rank per-subject adapter (rank $r$, $2\,C\,r$ parameters) matches the full encoder across five backbones, so subject conditioning earns its place on parameter cost rather than accuracy.}

\section{Methodology}

\subsection{EEG decoding notation}
\label{sec:decoding}

Let $\mathcal{S}=\{(\mathbf{X}_i,y_i,s_i)\}_{i=1}^{N}$ denote the sampled EEG dataset.
Each EEG trial $\mathbf{X}_i \in \mathbb{R}^{C \times T}$ is a multichannel epoch recorded from $C$ electrodes over $T$ time samples, $y_i \in \mathcal{Y}$ is its class label, and $s_i \in \{1,\ldots,K\}$ is its subject index.
For subject $k$, we write the subject-specific sample as
\[
\mathcal{S}_{k}=\{(\mathbf{X}_{i}^{(k)},y_{i}^{(k)})\}_{i=1}^{n_k}.
\]
All experiments are binary motor-imagery or motor-execution decoding problems with $\mathcal{Y}=\{0,1\}$; the concrete event pair is dataset-dependent and is specified in the dataset paragraph below.

EEG decoding can be formulated as a supervised classification problem: learn a mapping $f:\mathcal{X}\rightarrow\mathcal{Y}$ that associates brain activity with the corresponding mental state~\cite{King:encodingdecoding:2020}.

Within a deep learning framework, this mapping is approximated by a parametric model $f_{\theta}$.
Given a training split $\mathcal{S}_{\mathrm{tr}}$, its parameters are optimized by minimizing the empirical risk
\begin{equation}
\label{eq:empirical_risk}
\widehat{\mathcal{R}}(\theta)
  = \frac{1}{|\mathcal{S}_{\mathrm{tr}}|}
    \sum_{(\mathbf{X}_i,y_i)\in \mathcal{S}_{\mathrm{tr}}}
    \ell\!\left(f_{\theta}(\mathbf{X}_i),y_i\right),
\end{equation}
where $\ell$ is the supervised classification loss. \revised{Throughout this work, $\ell$ is the cross-entropy between the predicted class distribution and the one-hot label,}
\begin{equation}
\revised{\ell\!\left(f_{\theta}(\mathbf{X}_i), y_i\right) = -\sum_{c \in \mathcal{Y}} \mathbb{1}\!\left[y_i = c\right] \log p_{\theta}\!\left(c \mid \mathbf{X}_i\right),}
\end{equation}
\revised{where $p_{\theta}(\cdot \mid \mathbf{X}_i)$ is the softmax over the two output logits.}
We use this objective before adding subject-specific parameters.

End-to-end EEG classifiers often mimic signal-processing pipelines by separating representation extraction from classification~\cite{wimpff2024eeg, Lawhern2018, Chen2024EEGNeX}. We write the encoder as

\begin{equation}
e_{\phi}:\mathcal{X}\rightarrow\mathcal{Z},
\end{equation}

which maps a raw EEG trial to a latent representation $z_i=e_{\phi}(\mathbf{X}_i)\in\mathcal{Z}$. The classifier is

\begin{equation}
h_{\psi}:\mathcal{Z}\rightarrow\mathcal{Y},
\end{equation}

and the full model is

\begin{equation}
f_{\theta}=h_{\psi} \circ e_{\phi}.
\end{equation}

During training, feature extraction and classification are optimized together, so the objective can preserve task-relevant structure that a manual feature pipeline may discard~\cite{LIAN2024108727}.

\subsection{Subjects as domains}
\label{sec:adaptation}

Following the definition in~\cite{zhuang2020comprehensive}, a \emph{domain} consists of a feature space and a marginal distribution over that space, while a \emph{task} consists of a label space and the predictive function to be learned.
We denote by $\mathcal{S}_k$ the finite sample set of subject $k$ and the domain by
\[
\Omega_k = \bigl(\mathcal{X}, P_k(\mathbf{X})\bigr).
\]
The associated subject-level prediction task is described by $\tau_k=(\mathcal{Y}, q_k(y\mid\mathbf{X}))$.

In transfer-learning terms, each leave-one-subject-out fold selects one target subject $t$ and treats all other subjects as sources.
We denote the subject index set by $\mathcal{I}=\{1,\ldots,K\}$ and the source set for target $t$ by $\mathcal{I}_{-t}=\mathcal{I}\setminus\{t\}$.
The source data are $\mathcal{S}_{-t}=\bigcup_{k\in\mathcal{I}_{-t}}\mathcal{S}_k$, while the target data $\mathcal{S}_{t}$ are held out from model pre-training.
The goal is to use the labelled source domains $\{\Omega_k:k\in\mathcal{I}_{-t}\}$ to learn a model that transfers to $\Omega_t$ with only a small labelled calibration set from the target subject.

Traditional EEG decoding often assumes that training and testing samples are drawn from the same domain~\cite{Pan2010TL}. EEG signals violate this assumption because physiology, electrode placement, and non-stationarity vary across subjects~\cite{Lotte2018update}. 

In a single study, $\mathcal{X}$ and $\mathcal{Y}$ are typically shared across subjects, but the distributions differ. We distinguish two discrepancies:

\begin{itemize}
    \item \textbf{Covariate shift:} the marginal distributions differ, $P_i(\mathbf{X}) \neq P_k(\mathbf{X})$ for subjects $i \neq k$, even when the task remains fixed.
    \item \textbf{Concept shift:} the conditional label distributions differ, $q_i(y\mid\mathbf{X}) \neq q_k(y\mid\mathbf{X})$, so the same observed pattern may map to different labels across subjects.
\end{itemize}


Under these conditions, a single shared encoder $e$ assumes that one input transformation fits every subject. This may fail when inter-subject variability is large\revised{, and neglecting domain variations might lead to \emph{negative transfer}. Formally, negative transfer occurs when using the source domains degrades performance on the target relative to using target data alone: writing $\varepsilon_t(\cdot)$ for the expected risk on domain $\Omega_t$, it is the condition $\varepsilon_t\!\left(f(\mathcal{S}_{-t} \cup \mathcal{S}_t)\right) > \varepsilon_t\!\left(f(\mathcal{S}_t)\right)$, so the transferred model is worse than the one that ignores the sources~\cite{wang2019negativetransfer}.}

\begin{figure}[!htbp]
    \centering
    \includegraphics[width=0.85\linewidth]{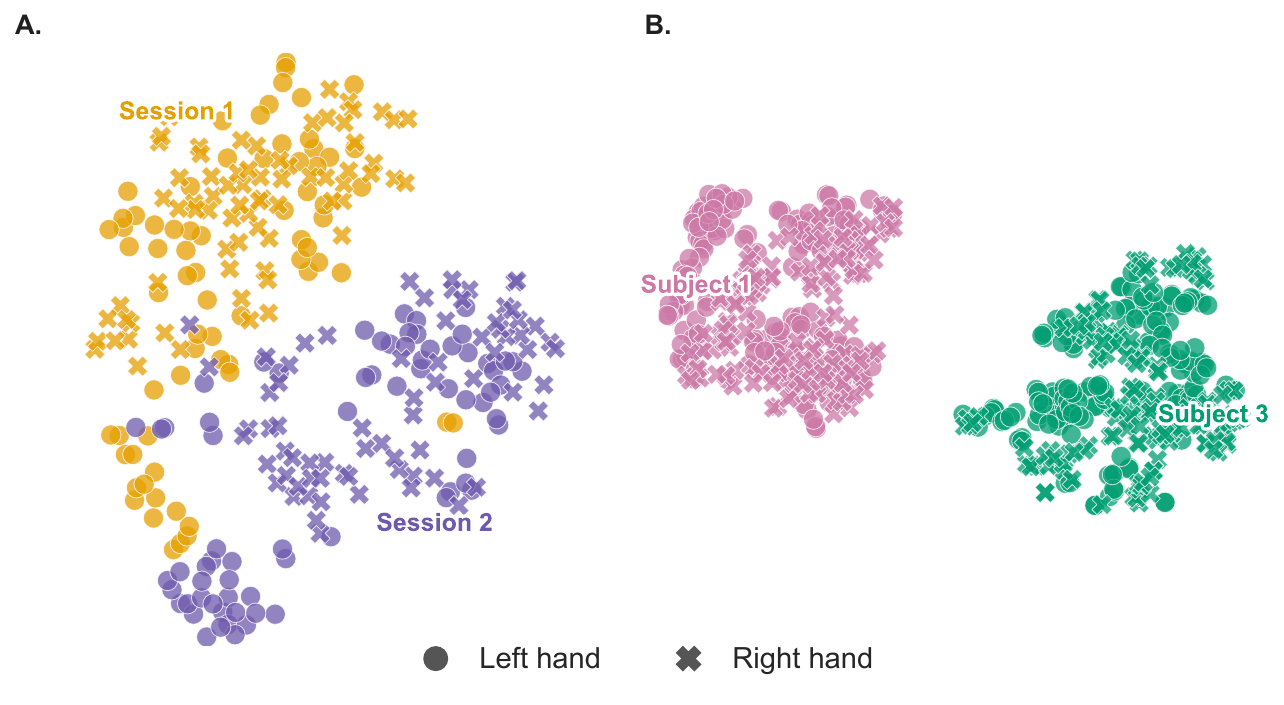}
    \caption{\textbf{The between-subject domain gap is larger than the within-subject one.}
    Two-dimensional projection of raw spatial-covariance features; marker shape encodes class (circle~=~left hand, cross~=~right hand). The two panels use different colour pairs because they compare different things.
    \textbf{(A)} Within-subject: the two sessions of one subject, coloured by the source/target role (orange source session~1 $\rightarrow$ purple target session~2). The sessions overlap heavily; the within-subject covariate shift, while non-zero, is small.
    \textbf{(B)} Between-subject: two different subjects (subject~1 and subject~3), each shown in its own subject colour. The two clouds are clearly separated, even though both subjects performed the same motor-imagery task with the same recording montage.
    This between-subject separation is the source of the negative-transfer problem motor-imagery decoders face when trained on pooled multi-subject data, and motivates both the alignment and architectural strategies developed in Sections~\ref{sec:adaptation}--\ref{sec:training_dynamics}.}
    \label{fig:domain_gap}
\end{figure}

Figure~\ref{fig:domain_gap} shows the scale of this gap: the between-subject shift is much larger than the within-subject, between-session shift, even though the task and recording protocol are the same.

\FloatBarrier

\subsection{Euclidean Alignment}

One strategy to mitigate inter-subject shifts is to apply transformations that align the distributions of EEG signals across domains. A widely used approach in EEG-based deep learning pipelines is \emph{Euclidean Alignment} (EA)~\cite{He2020:euclidean}.

EA aims to reduce covariate shift by recentering each subject with respect to its mean covariance matrix~\cite{Gretton2008:MMD}. Let $\mathbf{X}_i^{(k)} \in \mathbb{R}^{C \times T}$ denote the $i$-th EEG trial recorded from subject $k$, \revised{already temporally centered}. Each trial is represented by its spatial covariance matrix

\[
\Sigma_i^{(k)} = \frac{1}{T} \mathbf{X}_i^{(k)} {\mathbf{X}_i^{(k)}}^{\top}.
\]

EA then whitens each trial using the inverse square root of its \revised{euclidean} mean covariance matrix:

\begin{equation}
\widetilde{\mathbf{X}}_i^{(k)} = \bigl(\bar{\Sigma}^{(k)}\bigr)^{-1/2} \mathbf{X}_i^{(k)}
\end{equation}

The transformation is estimated independently for each \revised{run of each} subject and does not use class labels.

Covariance matrices belong to the space of symmetric positive definite (SPD) matrices~\cite{Riemannian2017,MDRM}. Under the affine-invariant Riemannian metric (AIRM) $\rho$, EA recentres each subject's covariance distribution around the identity:

\begin{equation}
\label{eq:ea}
\rho\!\left(\Sigma_i^{(k)}, \bar{\Sigma}^{(k)}\right)
= \rho\!\left(
\bigl(\bar{\Sigma}^{(k)}\bigr)^{-1/2}
\Sigma_i^{(k)}
\bigl(\bar{\Sigma}^{(k)}\bigr)^{-1/2},
I_C
\right).
\end{equation}

Equation~(\ref{eq:ea}) is written in covariance space, showing how the transformation acts in the distribution of covariances. Equation~\ref{eq:ea} shows how the same whitening matrix can be applied directly to the EEG trials. EA can therefore be inserted before a deep EEG model.

\revised{As already discussed by~\cite{He2020:euclidean}, whitening a set of trials using their own reference is an \emph{offline} strategy, not tuned for in-time processing. Therefore, in this work, we adapt this idea for \emph{online} settings by estimating the reference using the set of just the first $L = 24$ consecutive trials of subject $k$, defined as $\mathcal{R}^{(k)}$. The reference for that run is the Euclidean mean}

\begin{equation}
\revised{\bar{\Sigma}^{(k)} = \frac{1}{|\mathcal{R}^{(k)}|} \sum_{\Sigma_i \in \mathcal{R}^{(k)}} \Sigma_i^{(k)} ,}
\end{equation}

\revised{which is then used for recentering the rest of the subject, so that the alignment never accesses held-out trials. 
We selected 24 trials to match the length of a single run in the \textit{BNCI2014} dataset and by following the study conducted by \cite{martin_online:2024}, which recommends using between 16 and 32 trials for alignment.
}

In the experiments below, EA is therefore a subject-level preprocessing condition: the \emph{EA} runs use $\widetilde{\mathbf{X}}_i^{(k)}$, while the \emph{NoEA} runs use the same preprocessed trials before this whitening step.

\subsection{Multi-head architecture}
\label{sec:multihead}

Although Euclidean Alignment reduces inter-subject variability, it only recenters the covariances, and may leave other subject-specific patterns unchanged~\cite{rodrigues18:procrutes}.

To address this, we explore whether the architecture itself can learn subject-specific transformations during end-to-end training. Inspired by the standard \emph{feature extractor-classifier} model structure commonly used in EEG decoding~\cite{Lawhern2018, wimpff2024motor}, we replace the shared feature extraction encoder with a set of subject-specific encoders, while keeping a single shared classifier:
\[
f_{\theta}^{(k)}(\mathbf{X}) = h_{\psi}\!\left(e_{\phi_k}^{(k)}(\mathbf{X})\right),
\]
where \(e_{\phi_k}^{(k)}\) is the encoder specialized for subject \(k\), and \(h_{\psi}\) is the classifier shared across all subjects. Figure~\ref{fig:heads} shows a diagram of the proposed structure.

For a target fold $t$, the trainable hybrid parameter set is
\[
\Theta_{-t}=\left\{\psi,\{\phi_k:k\in\mathcal{I}_{-t}\}\right\}.
\]
The shared baseline is recovered by replacing the encoder bank with a single encoder $e_{\phi}$ used for every source subject. 


Within this architecture, each encoder can learn nonlinear subject-specific transformations of the raw EEG. The shared classifier keeps those representations compatible with one decision boundary, so the classification loss can act as an implicit alignment signal.

\revised{The \emph{capacity} of the per-subject encoder $e_{\phi_k}^{(k)}$ is itself a design choice. At full capacity it is a complete backbone, the hybrid described above. At its minimum capacity it reduces to a single per-subject \emph{low-rank layer}, following the line of~\cite{klein2025mitigating}: a rank-$r$ residual map applied to the raw trial,
\[
A_k(\mathbf{X}) = \mathbf{X} + U_k D_k \mathbf{X},
\qquad D_k \in \mathbb{R}^{r \times C},\quad U_k \in \mathbb{R}^{C \times r},
\]
where $C$ is the number of EEG channels and $r \le C$ is the rank. The matrix $D_k$ projects the $C$ channels onto an $r$-dimensional subject subspace and $U_k$ maps that subspace back to the channels, so $U_k D_k$ works as a correction specific to subject $k$. We initialise $U_k$ to zero, so $A_k$ starts as the identity and departs from it only as the subject-specific term is learned. The transformed trial then feeds the shared backbone $e_{\phi}$ and the shared classifier $h_{\psi}$, and the only per-subject parameters are the $2\,C\,r$ entries of $D_k$ and $U_k$, with the rank $r$ as the sole capacity knob. The full backbone and the rank-$r$ layer are therefore the same subject-specific encoder at two capacities; we denote them \texttt{+hybrid} and \texttt{+adapter} and, in the low-rank case, sweep $r \in \{1,2,4,8,16\}$. Both are trained and evaluated identically (Section~\ref{sub:train}), so they differ only in this per-subject module.}

Using this framework, we investigate four main questions:
\begin{enumerate}
    \item Can subject-specific encoders capture characteristic subject-dependent patterns in EEG signals?
    \item Does the multi-head structure produce latent representations that are better aligned across subjects?
    \item How do these representational changes affect cross-subject decoding performance?
    \item How does the proposed structure compare to the original backbone architectures?
\end{enumerate}

\subsection{Training and target-subject adaptation}
\label{sub:train}

Model training is divided into two stages: pre-training and fine-tuning.

During pre-training, each optimization step samples one mini-batch $\mathcal{B}_k$ per source subject $k\in\mathcal{I}_{-t}$ and routes it through the corresponding subject-specific encoder, as shown in Figure~\ref{fig:heads}.
The training objective for one target fold is
\begin{equation}
\label{eq:hybrid_training}
\min_{\Theta_{-t}}
\frac{1}{|\mathcal{I}_{-t}|}
\sum_{k\in\mathcal{I}_{-t}}
\frac{1}{|\mathcal{B}_k|}
\sum_{(\mathbf{X},y)\in\mathcal{B}_k}
\ell\!\left(h_{\psi}(e_{\phi_k}^{(k)}(\mathbf{X})),y\right).
\end{equation}
Gradients from all subject losses accumulate in the shared classifier, while each encoder receives gradients only from its own subject.
We update the model only after all subject batches have been processed.
The delayed update balances the shared classifier gradient across subjects while each encoder specializes in one subject.

Source training yields one shared classifier and one encoder per source subject. The encoders can be reused for their own subjects or transferred to a new subject.

For a held-out target subject, we initialize a target encoder from one of the pretrained source encoders and combine it with the shared classifier.
Let $\mathcal{C}_t\subset\mathcal{S}_t$ be the first 48 labelled calibration trials of the target subject, taken from the start of the first session, and let $\mathcal{T}_t=\mathcal{S}_t\setminus\mathcal{C}_t$ be the held-out target evaluation trials.
\revised{We use 48 since it consists two recording runs for \emph{BNCI2014} dataset, and we empirically noted that its a sufficient amount of calibration data for other datasets also. No trial of $\mathcal{T}_t$ contributes to head selection or fine-tuning.}
We select the source head
\begin{equation}
\label{eq:head_selection}
k^{\star}(t)
  = \argmax_{k\in\mathcal{I}_{-t}}
    \operatorname{BA}_{\mathcal{C}_t}
    \!\left(h_{\psi}\circ e_{\phi_k}^{(k)}\right),
\end{equation}
where $\operatorname{BA}_{\mathcal{C}_t}$ is balanced accuracy on the calibration set.
The selected encoder initializes the target branch by copying its weights, $\phi_t\leftarrow\phi_{k^{\star}(t)}$.
This target branch is then fine-tuned on $\mathcal{C}_t$ and evaluated on $\mathcal{T}_t$.
This is a low-calibration setting: head selection uses only target calibration data, while the oracle head reported in the results is an upper bound.

\subsection{Models and data splitting}
\label{sec:models}

We use \emph{EEGNet}~\cite{Lawhern2018}, \emph{AttentionBaseNet (AttnBNet)}~\cite{wimpff2024motor}, and \emph{CTNet}~\cite{zhao2024ctnet} as backbones because all three are compact EEG decoders used for motor imagery, spanning a purely convolutional model, an attention-augmented convolutional model, and a convolutional-transformer model.

EEGNet is a convolutional neural network (CNN) architecture for EEG classification. It uses depthwise and separable convolutions to mimic traditional EEG feature extraction, such as spatial filtering and filter-bank construction, before the final classification layer.

AttentionBaseNet is a compact attention-based backbone for motor-imagery EEG decoding that combines learned temporal filters with an attention module before classification.

CTNet is a convolutional-transformer backbone that uses an EEGNet-style convolutional stem to extract local spatial-temporal features, followed by a multi-head self-attention encoder that models global dependencies before classification.

\revised{This multi-head structure is a backbone-independent technical adaptation. Concretely, each backbone is split at a fixed depth: the modules before the cut are instantiated once per subject and form $e_{\phi_k}^{(k)}$, while the modules from the cut onwards are instantiated once and shared, and together with the final classification layer they form $h_{\psi}$. 
For all the architectures used in this work, cut is placed after the convolutional and attentional stem, so what is duplicated per subject is the spatial and temporal filtering, and what is shared is the remaining processing and the decision layer. Based on the braindecode~1.1.0~\cite{braindecode}, EEGNet is cut after the Separable Convolution (layer $12$), AttentionBaseNet is cut after the ChannelAttention block (module $3$), and CTNet is cut after the Convolutional and Attentional residual connection.}

We compare each backbone in its original shared-encoder form, \revised{LoRA}, and in the hybrid extension, all with and without EA. 

\paragraph{Implementation details.}
The training recipe is fixed across all three backbones, so that accuracy differences reflect the architecture and the alignment rather than per-model tuning.
All models are optimized with Adam. \revised{Pre-training uses a learning rate of $5\times10^{-4}$, weight decay of $1\times10^{-3}$, a batch size of $64$ per source subject, and a maximum of $500$ epochs with early stopping after $150$ epochs without improvement. Target fine-tuning uses a learning rate of $1\times10^{-4}$, weight decay of $5\times10^{-4}$, a batch size of $12$, and a maximum of $300$ epochs with the same patience of $150$. In both stages $20\,\%$ of the source trials are held out as the validation split used for checkpoint selection and early stopping.}
\revised{The complete evaluation is repeated across five independent seeds ($\{2342, 1234, 3751, 5087, 8291\}$, reseeding both the pre-training and the target-adaptation stage) for the three lighter datasets; the heaviest dataset (Schirrmeister2017) is evaluated at a single seed for cost, stated where relevant. 
}
The models, training loop, and Euclidean Alignment are implemented with the use of PyTorch, MOABB\cite{moabb2018}, and braindecode\cite{braindecode}; the code to reproduce the experiments and figures is linked under Code availability.
This separates preprocessing alignment (EA versus NoEA) from architectural alignment (shared encoder versus subject-specific encoders).
\cut{For the mechanism analyses, we also include a center-loss hybrid variant when available.}

\begin{figure}[htbp]
\centering
\includegraphics[width=\columnwidth]{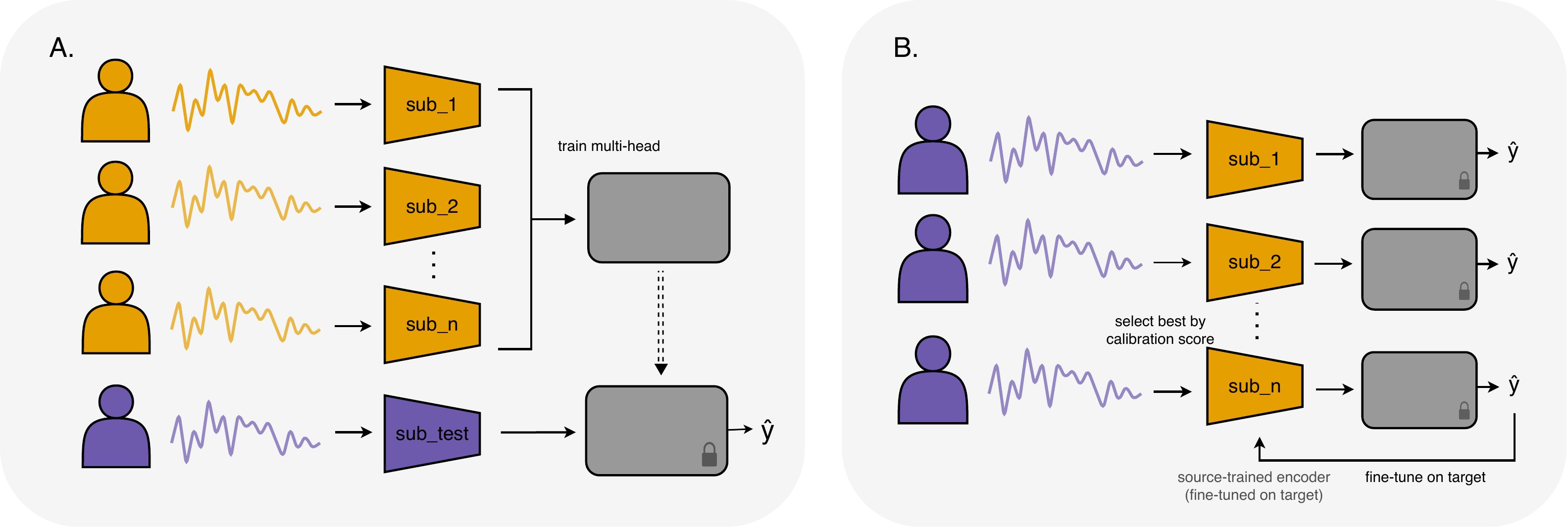}
\caption{\textbf{Overview of the proposed adaptation strategy for unseen subjects.} After training the multi-head model, each subject-specific encoder is combined with the shared classifier to form an individual initialization candidate. These candidate models are evaluated using calibration data from the target subject, and the best-performing initialization is selected for further fine-tuning.}
\label{fig:heads}
\end{figure}

Figure~\ref{fig:heads} illustrates the use of the trained subject-specific heads to initialize a new head for a target subject before fine-tuning.

\FloatBarrier

We train both shared and hybrid models with leave-one-subject-out (LOSO) splits.
For each target subject $t$, no trial from $\mathcal{S}_t$ is used during pre-training.
Within each source subject, we split trials into 80\% training and 20\% validation while preserving the subject assignment.
Model checkpoints are selected based on source validation performance; target calibration trials are used only after pre-training for head selection and fine-tuning.

\subsection{Datasets}
For our experiments, we used dataset IIa from BCI Competition 2014~\cite{BNCI2014001} (\emph{BNCI2014-001}), dataset I from BCI Competition 2015~\cite{BNCI2015} (\emph{BNCI2015-001}), High-Gamma~\cite{Schirrmeister2017} (\emph{Schirrmeister2017}), and the motor-imagery dataset from Yi et al.~\cite{Yi2014EEGMotorImagery} (\emph{Weibo2014}), all available through the Mother of All BCI Benchmarks (\textsc{MOABB})~\cite{Aristimunha_Mother_of_all_2023}.

\emph{BNCI2014-001} is a motor imagery (MI) dataset with four tasks (left hand, right hand, both feet, and tongue) and data from 9 healthy subjects, sampled on 22 channels and with a frequency of 250 Hz. The recordings are divided into two sessions on different days, with each session comprising six runs of 12 trials per class. We use the binary left-hand versus right-hand subset.
\emph{BNCI2015-001} is included as a second BCI Competition event-related desynchronization (ERD)/BCI benchmark. The MOABB version used here contains 12 subjects and two events, right hand and feet; we therefore use that binary task directly.

\emph{Schirrmeister2017} contains motor execution (ME) data with four classes (left hand, right hand, both feet, and rest) from 14 healthy subjects. It consists of two sessions with around 880 per subject in total, with variation across subjects.
It is divided into training and testing sessions and was originally sampled using 128 channels at 512 Hz. We use the binary left-hand versus right-hand subset. Subject 1 has only 240 trials, so we excluded it and limited the remaining subjects to the first 408 trials to keep batches balanced across subjects.

\emph{Weibo2014} contains data from 10 subjects and seven motor imagery conditions in MOABB: left hand, right hand, both hands, feet, compound hand-foot imagery, and rest. We use the same binary left-hand versus right-hand subset used for the other left/right datasets.

Apart from the \emph{Schirrmeister2017} trial balancing, we used standard motor-imagery preprocessing~\cite{nam2018brain, Aristimunha_Mother_of_all_2023}: $8$--$32\,\mathrm{Hz}$ band-pass filtering with overlap-add and resampling to $250\,\mathrm{Hz}$.
\revised{Channel selection differs across datasets and is dictated by the montage each one provides. \emph{BNCI2014-001} is used with its full $22$-channel montage~\cite{BNCI2014001}. For \emph{Schirrmeister2017} and \emph{Weibo2014} we restrict the recording to a common $21$-channel sensorimotor grid covering the FC, C, and CP rows (FC5--FC6, C5--C6, CP5--CP6, including the midline FCz, Cz, and CPz), which spans the sensorimotor cortex where the motor-imagery and motor-execution responses of interest are expressed. \emph{BNCI2015-001} is distributed with $13$ channels and is used as provided.}

\paragraph{Ethical approval} This paper does not contain any studies with human or animal participants performed by any of the authors.

\subsection{Evaluation metrics and representation diagnostics}
\label{sec:evaluation_diagnostics}

The primary decoding metric is balanced accuracy on the held-out target trials,
\begin{equation}
\label{eq:balanced_accuracy}
\operatorname{BA}
  = \frac{1}{|\mathcal{Y}|}
    \sum_{c\in\mathcal{Y}}
    \frac{\operatorname{TP}_{c}}
         {\operatorname{TP}_{c}+\operatorname{FN}_{c}},
\end{equation}
which weights the two classes equally and makes the score comparable across datasets and subjects.
For the hybrid model, \texttt{+hybrid} denotes the head selected by equation~(\ref{eq:head_selection}), whereas \texttt{+hybrid\textsuperscript{*}} denotes the oracle head selected by the target test score. The oracle is reported only to quantify how much performance remains available to a better non-oracle head-selection rule.
We summarise paired across-subject comparisons with the Wilcoxon signed-rank test, report Wilson score intervals on win-rates, and use the matched-pairs rank-biserial correlation $r$ as an effect size; within each family of comparisons, p-values are Holm-corrected for multiple testing.
We use non-parametric tests because the per-subject score differences are not assumed normal and the per-dataset samples are small.
Unless stated otherwise, tests are two-sided at a family-wise level of $\alpha = 0.05$ after Holm correction. Comparisons pooled across datasets use all $n = 44$ subjects ($9 + 12 + 13 + 10$); per-dataset comparisons use that dataset's subjects.

To analyse what the models learn beyond accuracy, we compute diagnostics on trial-level covariance matrices.
For a representation map $g$ (raw signal, shared encoder, or a subject-specific encoder) \revised{temporally centered}, the covariance of a trial $i$ is
\[
\Gamma_i^{g} = \frac{1}{T_g} g(\mathbf{X}_i)g(\mathbf{X}_i)^{\top},
\]
where $T_g$ is the temporal length of that representation.

Class distinctiveness \cite{lotte2018skills} extends the Fisher criterion to covariance matrices using the same affine-invariant Riemannian metric $\rho$ on the SPD manifold. For two classes $\mathcal{K}_{0}, \mathcal{K}_{1}$ with Riemannian means $M_{0}, M_{1}$ and intra-class dispersions $\sigma(\mathcal{K}_{0}), \sigma(\mathcal{K}_{1})$,
\begin{equation}
\label{eq:class_dist}
\operatorname{ClassDis}(\mathcal{K}_{0}, \mathcal{K}_{1})
  \;=\; \frac{\rho(M_{0}, M_{1})}
             {\tfrac{1}{2}\bigl(\sigma(\mathcal{K}_{0}) + \sigma(\mathcal{K}_{1})\bigr)},
\qquad
\sigma(\mathcal{K}_{c})
  \;=\; \frac{1}{|\mathcal{K}_{c}|} \sum_{\Sigma \in \mathcal{K}_{c}} \rho(\Sigma, M_{c}).
\end{equation}
Large values indicate classes whose centroids are far apart and whose internal spread is small.

\section{Results}
\label{sec:results}

\subsection{Euclidean Alignment recenters subjects and translates to per-subject accuracy gains}
\label{sec:ea_effect}

\begin{figure}[!htbp]
    \centering
    \includegraphics[width=\linewidth]{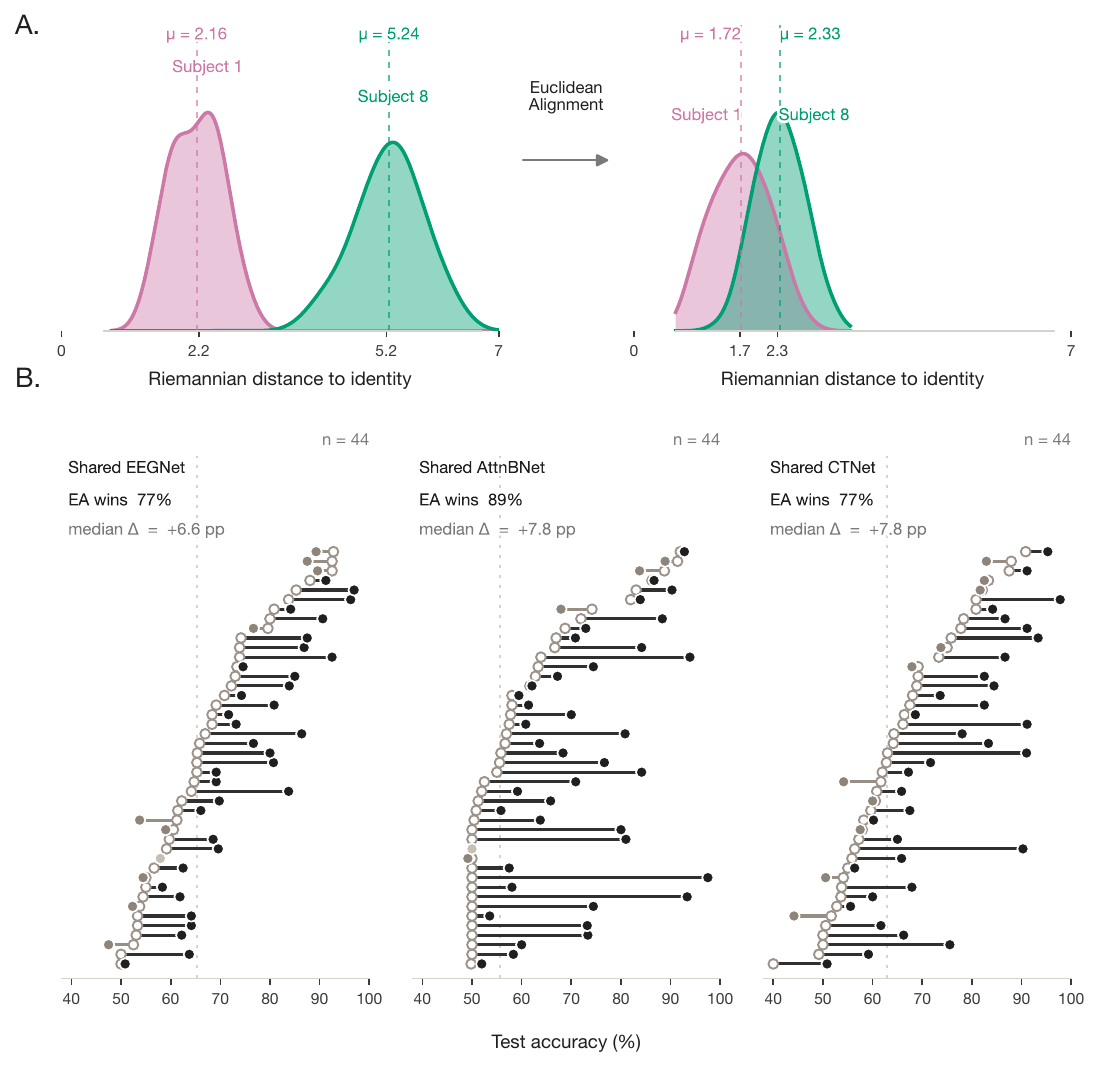}
    \caption{\textbf{Euclidean Alignment recenters each subject's covariance onto a shared reference, and this translates to per-subject accuracy gains.}
    \textbf{(A)} \emph{Mechanism.} Riemannian distance-to-identity for two simulated subjects ($n = 200$ trials each). Without EA, the subjects sit in separated distance regimes ($\mu_{S1} = 2.16$, $\mu_{S8} = 5.24$). EA whitens each subject's trials with the inverse square root of its own \revised{Euclidean mean covariance}; both distributions collapse onto a shared regime around the identity ($\mu_{S1} = 1.72$, $\mu_{S8} = 2.33$).
    \textbf{(B)} \emph{Effect.} Per-subject NoEA $\rightarrow$ EA test accuracy for the shared-encoder baseline on three backbones (EEGNet, AttnBNet, CTNet). Each row is one subject; the open grey circle marks NoEA and the filled circle marks EA, joined by a segment drawn in dark ink where EA improves accuracy ($\Delta_{\mathrm{EA-NoEA}} > 0$) and in grey where it does not. Rows are sorted by NoEA accuracy. EA wins on a majority of subjects on every backbone ($77\,\%$, $89\,\%$, $77\,\%$; Wilson $95\,\%$ CIs $63$--$87$, $76$--$95$, $63$--$87$), with median improvements of $+6.6$, $+7.8$, and $+7.8$ percentage points.}
    \label{fig:ea_mechanism_effect}
\end{figure}

We begin by separating the mechanism through which EA acts from the effect it has on real decoding. Figure~\ref{fig:ea_mechanism_effect} reports both. Panel A simulates two subjects whose covariances sit at different points on the SPD manifold; EA's whitening collapses their distributions to the same center.
Panel B then carries the same intervention to the empirical data: for the shared-encoder baseline on three backbones, we report the per-subject paired NoEA $\rightarrow$ EA test-accuracy delta. EA wins on the majority of subjects on every backbone, with median improvements of $+6.6$ to $+7.8$ percentage points (Wilcoxon signed-rank $p < 10^{-5}$ on every backbone, Holm-corrected; rank-biserial $r = 0.80$--$0.91$). The mechanism in A and the per-subject effect in B answer two distinct objections (\emph{does EA do something geometrically meaningful?} and \emph{does it transfer to subject-level accuracy?}) and form the baseline against which all later architectural changes are compared.

\FloatBarrier

\subsection{From preprocessing to architecture: making alignment learnable inside the model}
\label{sec:architecture_intro}

Figure~\ref{fig:ea_mechanism_effect} confirms that EA preprocessing yields per-subject accuracy gains, but EA is a fixed transformation applied separately to inputs. A natural follow-up question is whether subject-specific transformations can be \emph{learned} end-to-end as part of the model. We pursue this by replacing the shared feature encoder with the multi-head architecture of Section~\ref{sec:adaptation}: each subject's trials are routed through a dedicated encoder $e_{\phi_k}^{(k)}$ before reaching the shared classifier $h_{\psi}$. The schematic in Figure~\ref{fig:transformacao_no_modelo} shows the decomposition $f_{\theta}^{(k)}(\mathbf{X}) = h_{\psi}(e_{\phi_k}^{(k)}(\mathbf{X}))$ and the mechanism by how the per-subject encoders could learn some similar transformation by specializing on filtering individual patterns.

\begin{figure}[!htbp]
    \centering
    \definecolor{archSource}{HTML}{E69F00}
    \definecolor{archGrey}{HTML}{EEEAE3}
    \definecolor{archGreyLine}{HTML}{7A7A7A}
    \begin{tikzpicture}[
        font=\sffamily\small,
        node distance=0.72cm and 0.72cm,
        block/.style={draw, rounded corners=1.2pt, align=center, minimum height=0.78cm, inner xsep=0.16cm, fill=gray!7},
        shared/.style={block, fill=archGrey, draw=archGreyLine},
        subject/.style={block, fill=archSource!22, draw=archSource!75!black},
        latent/.style={block, draw=archGreyLine, fill=archGrey, minimum height=0.72cm, minimum width=0.72cm, inner sep=0pt},
        ghost/.style={dashed, opacity=0.45},
        arrow/.style={-Latex, thick}
    ]
    \node[align=left, font=\sffamily\bfseries] at (-0.2,1.25) {Fixed preprocessing};
    \node[block] (x1) at (0,0) {trial\\$\mathbf{X}^{(k)}$};
    \node[subject, ghost, right=of x1] (ea) {$\mathrm{EA}_k$\\$\bar{\Sigma}^{(k)}$};
    \node[block, ghost, right=of ea] (xt) {aligned trial\\$\widetilde{\mathbf{X}}^{(k)}$};
    \node[shared, right=of xt] (enc1) {shared encoder\\$e_{\phi}$};
    \node[latent, right=of enc1] (z1) {$z$};
    \node[shared, right=of z1] (clf1) {classifier\\$h_{\psi}$};
    \node[block, right=of clf1] (y1) {$\hat{y}$};
    \draw[arrow, ghost] (x1) -- (ea);
    \draw[arrow, ghost] (ea) -- (xt);
    \draw[arrow, ghost] (xt) -- (enc1);
    \draw[arrow] (enc1) -- (z1);
    \draw[arrow] (z1) -- (clf1);
    \draw[arrow] (clf1) -- (y1);

    \node[align=left, font=\sffamily\bfseries] at (-0.2,-1.55) {Learned subject-specific alignment};
    \node[block] (x2) at (0,-2.8) {trial\\$\mathbf{X}^{(k)}$};
    \node[subject, right=of x2, minimum width=2.65cm] (enc2) {subject-specific\\encoder\\$e_{\phi_k}^{(k)}$};
    \node[latent, right=of enc2] (z2) {$z^{(k)}$};
    \node[shared, right=of z2] (clf2) {shared classifier\\$h_{\psi}$};
    \node[block, right=of clf2] (y2) {$\hat{y}$};
    \draw[arrow] (x2) -- (enc2);
    \draw[arrow] (enc2) -- (z2);
    \draw[arrow] (z2) -- (clf2);
    \draw[arrow] (clf2) -- (y2);
    \node[align=left, font=\sffamily\scriptsize, text width=9.8cm] at (4.35,-4.0)
        {The hybrid model removes the explicit $\mathrm{EA}_k$ block and lets $e_{\phi_k}^{(k)}$ learn the subject-conditioned transformation needed by the shared classifier.};
    \end{tikzpicture}
    \caption{\textbf{The hybrid architecture replaces EA's fixed preprocessing step with a learned per-subject encoder.}
    Given a trial $\mathbf{X}$ from subject $k$, a subject-specific encoder $e_{\phi_k}^{(k)}$ maps the signal into a shared latent space, and a single classifier $h_{\psi}$ produces the prediction.
    This factorization lets the encoder learn a subject-conditioned transformation that plays a role similar to $\mathrm{EA}_k(\mathbf{X})$ through the supervised loss.}
    \label{fig:transformacao_no_modelo}
\end{figure}
\FloatBarrier

\subsection{Subject-specific gains over the plain EA baseline are dataset- and backbone-dependent}
\label{sec:methods_forest}

\begin{figure*}[!htbp]
    \centering
    \includegraphics[width=\linewidth]{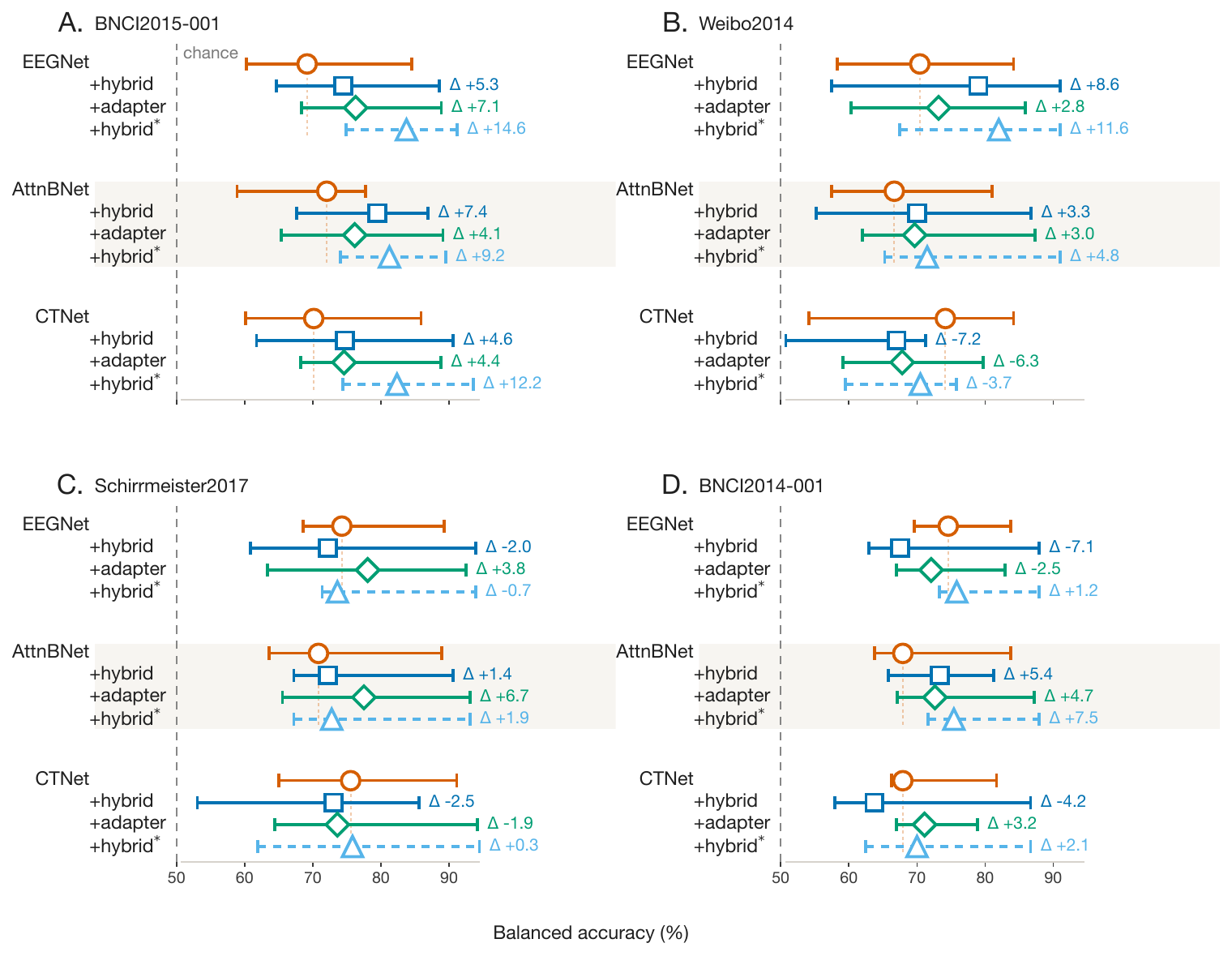}
    \caption{\textbf{Per-subject balanced accuracy of the proposed \texttt{+hybrid} extension, \revised{a low-rank per-subject adapter (\texttt{+adapter})}, and the oracle ceiling \texttt{+hybrid\textsuperscript{*}}, compared to plain baselines, across three backbones (\texttt{EEGNet}, \texttt{AttnBNet}, \texttt{CTNet}) on four motor-imagery datasets under Euclidean Alignment.}
Open markers give the subject median; horizontal bars span the inter-quartile range; the dashed line marks chance accuracy. \revised{Within each backbone block the rows are, top to bottom, the plain baseline (vermillion circle), the full per-subject encoder \texttt{+hybrid} (blue square), the rank-$2$ low-rank adapter \texttt{+adapter} (green diamond), and the oracle \texttt{+hybrid\textsuperscript{*}} (light-blue triangle). The adapter is deployable and calibration-selected like \texttt{+hybrid}, differing only in the per-subject module.} The thin vermillion vertical guide marks that block's plain-baseline EA median; the $\Delta$ annotation on each \texttt{+hybrid}, \revised{\texttt{+adapter},} and \texttt{+hybrid\textsuperscript{*}} row reports median balanced accuracy minus that baseline median, both measured under EA. The oracle variant selects each subject's best test-set head and therefore serves only as an upper bound for any validation-based head selection. Datasets are arranged in reading order (top-left to bottom-right) by the average \texttt{+hybrid} $-$ baseline gap across the three backbones, so the largest hybrid wins appear in the top-left panel.}
    \label{fig:rank_of_models}
\end{figure*}

Before turning to the mechanism, we report the cross-dataset performance results. Figure~\ref{fig:rank_of_models} compares each plain backbone trained under Euclidean Alignment (\texttt{EEGNet}, \texttt{AttnBNet}, or \texttt{CTNet}~\cite{zhao2024ctnet} + EA) against its hybrid extension on four motor-imagery datasets, and against an oracle ceiling \texttt{+hybrid\textsuperscript{*}} that selects each subject's best test-set head (an upper bound, not a deployable policy). The vertical guide within each backbone block marks that block's plain-baseline EA median, so $\Delta$ values read directly as the hybrid lift over the same backbone trained without subject-specific heads.

We can observe that the lift is not uniform across datasets or backbones. BNCI2015-001 is the one dataset where the hybrid model wins on all three backbones ($\Delta = +5.3$, $+7.4$, $+4.6$ pp for EEGNet, AttnBNet, CTNet), and the oracle ceiling lifts each further (to $+14.6$, $+9.2$, $+12.2$ pp). The other three datasets are mixed. On Weibo2014 the EEGNet and AttnBNet hybrids gain ($+8.6$ and $+3.3$ pp) but the CTNet hybrid collapses ($-7.2$ pp) with an oracle that stays below baseline ($-3.7$ pp), indicating a genuine failure of the CTNet hybrid representation on this dataset rather than a head-selection artefact. On Schirrmeister2017 the hybrid model is roughly flat on every backbone ($-2.5$ to $+1.4$ pp), and the oracle only recovers to about baseline. On BNCI2014-001 the EEGNet and CTNet hybrids underperform their plain baselines ($-7.1$ and $-4.2$ pp) while AttnBNet gains ($+5.4$ pp); for both losing backbones the oracle nearly closes the gap ($+1.2$ and $+2.1$ pp), implicating validation-driven head selection rather than the representation itself. The per-cell $\Delta$ values are a descriptive map of where the lift concentrates rather than independent confirmatory tests; the confirmatory comparison pools subjects within each backbone across the four datasets (one paired test per backbone, Holm-corrected across the three). This split is significant only for AttnBNet, whose hybrid improves over its EA baseline (median $+1.9$ pp, Wilcoxon $p = 0.014$, Holm $p = 0.042$); the EEGNet hybrid is indistinguishable from baseline ($+1.2$ pp, $p = 0.26$) and the CTNet hybrid trends downward ($-4.2$ pp, $p = 0.06$). The remaining sections decompose where the lift comes from (class distinctiveness, head-subject specificity, latent geometry) and where it breaks (subject heterogeneity, head-selection failure modes, and backbone-specific collapse). 

\FloatBarrier

\revised{The subject-specific encoder can be realized at very different capacities, and its capacity barely changes what it achieves: where a subject-specific extension helps at all, the full per-subject encoder and its low-rank form recover the same gain. Reducing each subject's encoder to the low-rank layer of Section~\ref{sec:multihead}, under the same leave-one-subject-out protocol and calibration-selected head and sweeping the rank $r \in \{1,2,4,8,16\}$, matched or exceeded the full per-subject encoder on every backbone, as highlighted by Table~\ref{tab:lora_rank}. At rank $r{=}1$ the per-subject cost was $2\,C\,r$ parameters, of order $10$, against $10^{3}$ to $10^{5}$ for a full encoder. This is evidence that subject-specific contribution can be well modeled as a low-rank residual addition to a main, common pattern rather than a full encoder bank.}

\begin{table}[htbp]\centering
\caption{\revised{Deployable balanced accuracy (\%) of a per-subject low-rank adapter (rank $r$) versus the full per-subject encoder (Hybrid), for the same backbone, target-adaptation, and calibration-selected head; prospective Euclidean Alignment, mean over the four datasets. The per-subject cost is $2\,C\,r$ parameters ($C = 13$--$22$ depending on the dataset montage): tens of parameters at $r{=}1$, against $10^3$--$10^5$ for a full encoder. On every backbone the adapter matches or beats the full encoder.}}
\label{tab:lora_rank}
\begin{tabular}{lcccccc}
\toprule
Backbone & Hybrid & $r{=}1$ & $r{=}2$ & $r{=}4$ & $r{=}8$ & $r{=}16$ \\
\midrule
EEGNet           & $71.6$ & $73.6$ & $75.9$ & $\mathbf{76.2}$ & $76.0$ & $74.2$ \\
CTNet            & $70.5$ & $74.5$ & $74.1$ & $74.6$ & $74.7$ & $\mathbf{75.4}$ \\
AttentionBaseNet & $72.8$ & $75.5$ & $\mathbf{75.6}$ & $75.5$ & $74.9$ & $75.1$ \\
\bottomrule
\end{tabular}
\end{table}

\subsection{Class distinctiveness and the role of subject-specific encoders}
\label{sec:class_distinctiveness}

A model can only classify what its representations separate.
Classification accuracy alone confounds two distinct competences: how well the encoder produces \emph{distinct} class patterns, and how well the downstream classifier exploits them~\cite{lotte2018skills}.
To isolate the first, we use the class-distinctiveness diagnostic introduced in equation~(\ref{eq:class_dist}), a classifier-independent quantity computed directly on the trial-level covariances of the learned representations.

Two properties make this metric well-suited to our diagnostic.
First, $\rho$ is affine-invariant, so $\operatorname{ClassDis}$ does not depend on the linear classifier head; it scores the representation, not the decision boundary.
Second, the metric is computed inside each subject's own trials, so the geometry of the latent space is isolated from any cross-subject alignment effect (which Figure~\ref{fig:ea_mechanism_effect} already accounts for separately).

\begin{figure}[!htbp]
    \centering
    \includegraphics[width=\linewidth]{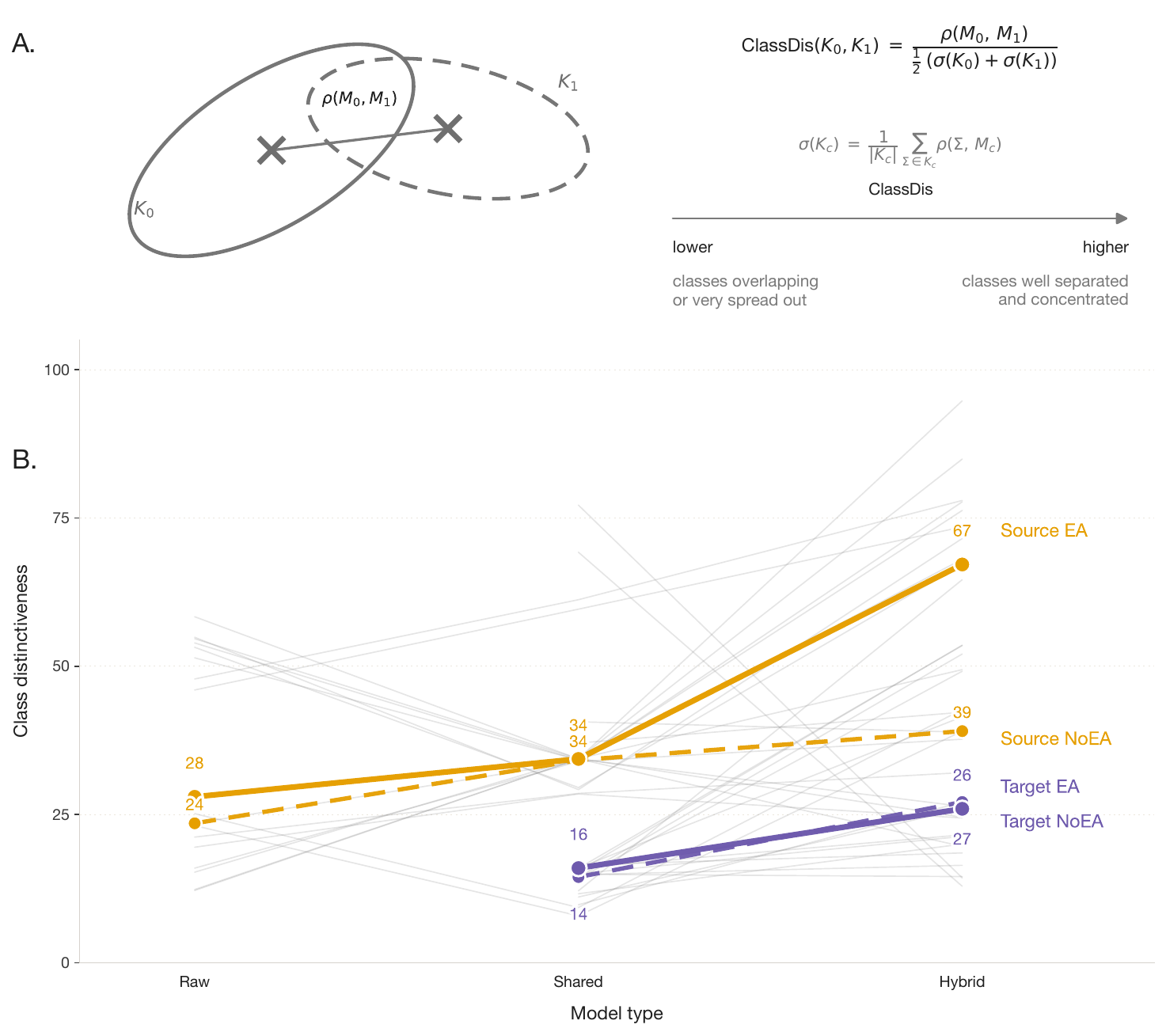}
    \caption{\textbf{Subject-specific encoders sharply lift class distinctiveness for source subjects, with a smaller gain that still generalises to held-out target subjects.}
    \textbf{(A)} The metric compares inter-class Riemannian distance to average intra-class dispersion.
    \textbf{(B)} Median per-subject distinctiveness across the pipeline (Raw, Shared, Hybrid) for source subjects (orange) and held-out target subjects (purple), with Euclidean Alignment (solid) and without (dashed); faint lines are individual subjects. The hybrid encoder lifts source distinctiveness well above the raw signal and the shared encoder ($28 \rightarrow 67$ under EA); for unseen target subjects the gain is smaller ($16 \rightarrow 26$) but stays above the shared-encoder level, indicating the subject-specific encoders retain transferable structure out of domain.}
    \label{fig:class_distinctiveness_concept_effect}
\end{figure}

We compute $\operatorname{ClassDis}$ on three representations of the same trials (Figure~\ref{fig:class_distinctiveness_concept_effect}B): the raw spatial covariance; the latent covariance after the shared encoder; and the latent covariance after the hybrid encoder. Since the analysis of source and target separation would reveal different things about the model, such as how much each encoder learned about their subject vs how much this can generalize to different subjects, we divided the analysis into source and target distinctiveness. For the target on the hybrid model, we didn't use the fine-tuned encoders as we wanted to assess the generalization on the pre-trained versions.

From the figure, we observe that the shared encoder slightly increases the distinctiveness of each source subject's embeddings relative to the raw signal ($34$ vs $28$ under EA; $34$ vs $24$ without). For the source subjects, both aligned and non-aligned models have approximately the same average class distinctiveness, which could make sense, since EA is simply a translation and shouldn't affect class separation; this is the role of feature extraction to select features that better separate them. For held-out target subjects, however, that value decreases to $16$ under EA vs $14$ , \revised{with no adaptation}, which could indicate the generalization of the embedding extraction capability of the encoder is still limited.

The hybrid encoder, however, sharply increases it for source subjects routed through their own head ($67$ under EA, $39$ without), which could indicate a high level of specialization. The difference between the hybrid EA and the regime without EA could be interpreted as additional improvement transformations: the model does not have to deal with aligning covariances, and thus can devote its capacity to non-linear transformations that further improve class separability. For held-out target subjects, distinctiveness falls, yet remains higher than under the shared encoder in both conditions, indicating that subject-specific encoders retain some transferable structure even when applied out of domain.

Two cautions apply to this diagnosis.
$\operatorname{ClassDis}$ scores the trial-level spatial covariance of the representation, so any separability that lives only in a non-linear classifier head downstream of the covariance read is invisible to it.
The metric also requires SPD covariances; if an encoder produces rank-deficient features (e.g., when an output channel is constant across trials), the Riemannian distance is undefined, and we restricted the analysis to representations where this did not occur.

\FloatBarrier

This metric measures an encoder-related quantity. Thus, an important question is whether \emph{ $\operatorname{ClassDis}$ acts as a diagnostic that relates to classifier behaviour, or is solely a geometric quantity that drifts from what the classifier cares about?} 

\begin{figure}[!htbp]
    \centering
    \includegraphics[width=0.72\linewidth]{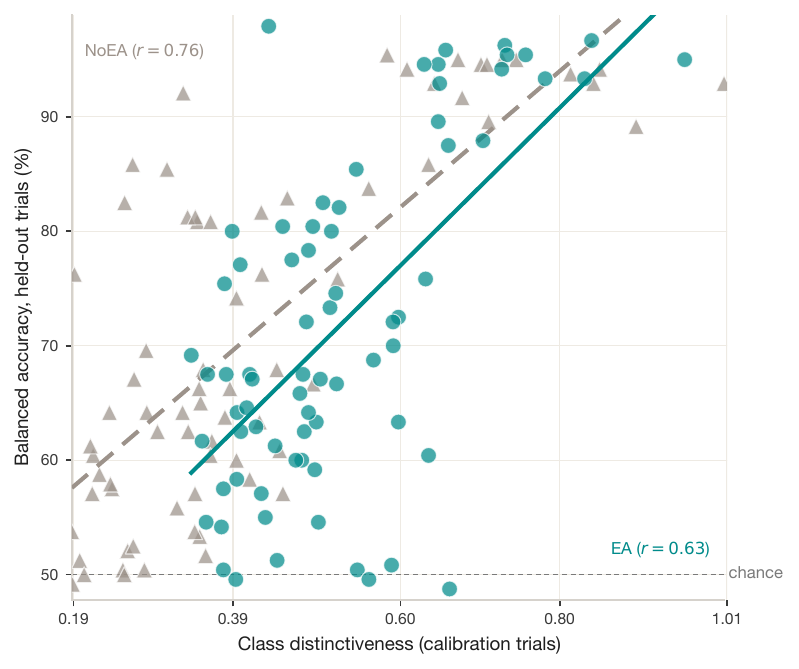}
    \caption{\textbf{Class distinctiveness predicts downstream classification accuracy, under both alignments.}
    Each point is one (target subject, encoder head) pair (NoEA: grey triangles; EA: teal circles); lines are least-squares fits (grey dashed = without EA, teal solid = with EA).
    Pooled across pairs the correlation is strong under both conditions (Pearson \revised{$r = 0.76$} without EA, \revised{$r = 0.63$} with EA), showing that the affine-invariant, classifier-free $\operatorname{ClassDis}$ score tracks balanced accuracy rather than being an artefact of the alignment step.}
    \label{fig:classdis_vs_score}
\end{figure}

Figure~\ref{fig:classdis_vs_score} plots per-pair $\operatorname{ClassDis}$ against balanced accuracy under both alignments. \revised{In the case where the two quantities are both computed from the same labelled trials and the same learned representations, a positive association is expected regardless of whether 
$\operatorname{ClassDis}$ carries any prospective information.
To isolate this possibility, we instead compute $\operatorname{ClassDis}$ on the \emph{calibration} trials and correlate it with balanced accuracy on the held-out \emph{evaluation} trials from that same encoder head. 
This calibration-to-evaluation correlation remains strong (Pearson $r=0.76$ without EA, $r=0.63$ with EA), indicating that $\operatorname{ClassDis}$ prospectively predicts decoding performance on unseen data and could support non-oracle head selection. Moreover}, these results highlight that $\operatorname{ClassDis}$ could act as a proxy for decoding accuracy under both alignments, which justifies using it for the EA-versus-NoEA comparisons that follow.

\FloatBarrier

\subsection{Each hybrid head learns its subject's manifold}
\label{sec:head_subject_specificity}

Class distinctiveness (Section~\ref{sec:class_distinctiveness}) confirms that the hybrid encoder produces more separable representations, especially for their own subjects, but it does not yet show that each head genuinely \emph{prefers} them. The control we report in Figure~\ref{fig:class_distinctiveness_concept_effect}B already pointed in that direction: passing a subject's trials through a different subject's head sharply decreases class distinctiveness values. We now make the head-subject geometry explicit on the manifold itself.

For each target subject we compute two normalized Riemannian distances on the trial-level latent covariances: the distance from the target's trials to their own head's centroid (\emph{target to head}), and the distance from the same trials to every other head's centroid (\emph{target to others}). If the heads are truly subject-specific, the former should be small and the latter large; if the heads converged to a common transformation, the two would be indistinguishable.

\begin{figure}[!htbp]
    \centering
    \includegraphics[width=\linewidth]{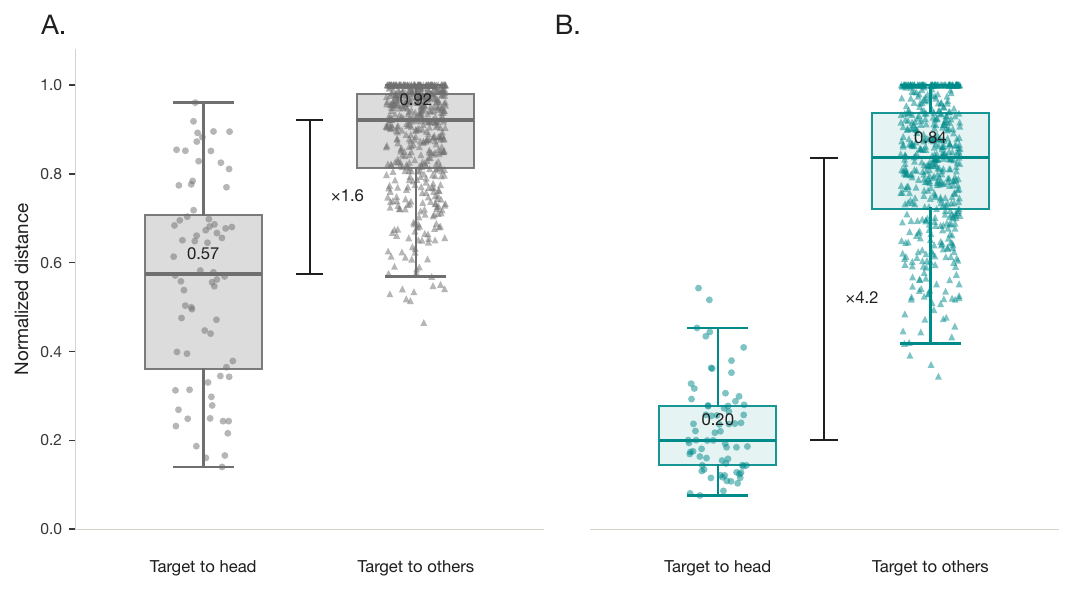}
    \caption{\textbf{Each subject's trials sit closer to their matched head's manifold than to other heads', and the gap widens under Euclidean Alignment.}
    Normalized Riemannian distance from target-subject trials to per-head encoder centroids in the hybrid model's latent covariance space, pooled over the leave-one-subject-out folds. In each panel the left box (circle points) is the distance to the matched head's own subject (\emph{target-to-head}); the right box (triangle points) is the distance to the other heads' subjects (\emph{target-to-others}).
    \textbf{Left (without EA).} The two distributions overlap, with a median gap of $\approx 1.6\times$; the panel is drawn in neutral grey.
    \textbf{Right (with EA).} Target-to-head drops well below target-to-others, a median gap of $\approx 4.2\times$; the panel is drawn in teal to mark the aligned condition.
    Box edges are the inter-quartile range; whiskers extend to $1.5\,$IQR; points are subject-head pairs.}
    \label{fig:dist_target_head}
\end{figure}
\FloatBarrier

Separation is present under both alignment regimes but is far larger after EA ($\approx 4.2\times$ versus $\approx 1.6\times$).
\revised{The $72$ matched and $504$ mismatched subject-head pairs are not independent: they are generated by only nine leave-one-subject-out folds, and within a fold the same eight encoders and eight source subjects recur in every cell. We therefore test also the effect at the level of the fold, which is the unit of replication. For each fold, we take the contrast between the mean matched distance and the mean mismatched distance, giving nine independent values. The matched head is closer in every fold under both alignments ($9/9$; median contrast $-0.36$ without EA and $-0.61$ with EA; one-sided Wilcoxon signed-rank $p = 1.95\times10^{-3}$ in both conditions).} Once EA places the source and target distributions in a common centre, each subject's trials are filtered most cleanly by their own head, because the encoder is trained to select the features most relevant for that subject; a mismatched head projects the same trials farther away.

\revised{A shared encoder widened to the hybrid's total parameter count neither beats nor loses to the normal shared (differences $\le 2.4$ pp, sign varying across datasets, within between-subject variability); and a subject-specific batch-normalisation baseline matches the hybrid under Euclidean Alignment but loses by $6.6$ pp on average without it. Thus, the effect could be attributable to subject-specific structure rather than to capacity, head multiplicity, or normalisation alone.} 
This also explains the class-distinctiveness decrease under the mismatched head situation in Figure~\ref{fig:class_distinctiveness_concept_effect}B, and sets up the next question: if each head encodes its own subject, what does the resulting cross-subject geometry look like at the architecture level?

\subsection{The hybrid encoder internalizes EA's alignment role}
\label{sec:hybrid_absorbs_ea}

Two independent measurements point in the same direction: the hybrid architecture's per-subject encoders take over the functional role that EA plays externally on the shared model, reaching a similar effect through learned feature transformations rather than an explicit covariance recentring. We report the evidence from two angles: training dynamics (validation-loss convergence with and without EA) and latent geometry (pairwise inter-subject distance in the learned representation).

\paragraph{Training dynamics.} \label{sec:training_dynamics}

\begin{figure}[!htbp]
    \centering
    \includegraphics[width=\linewidth]{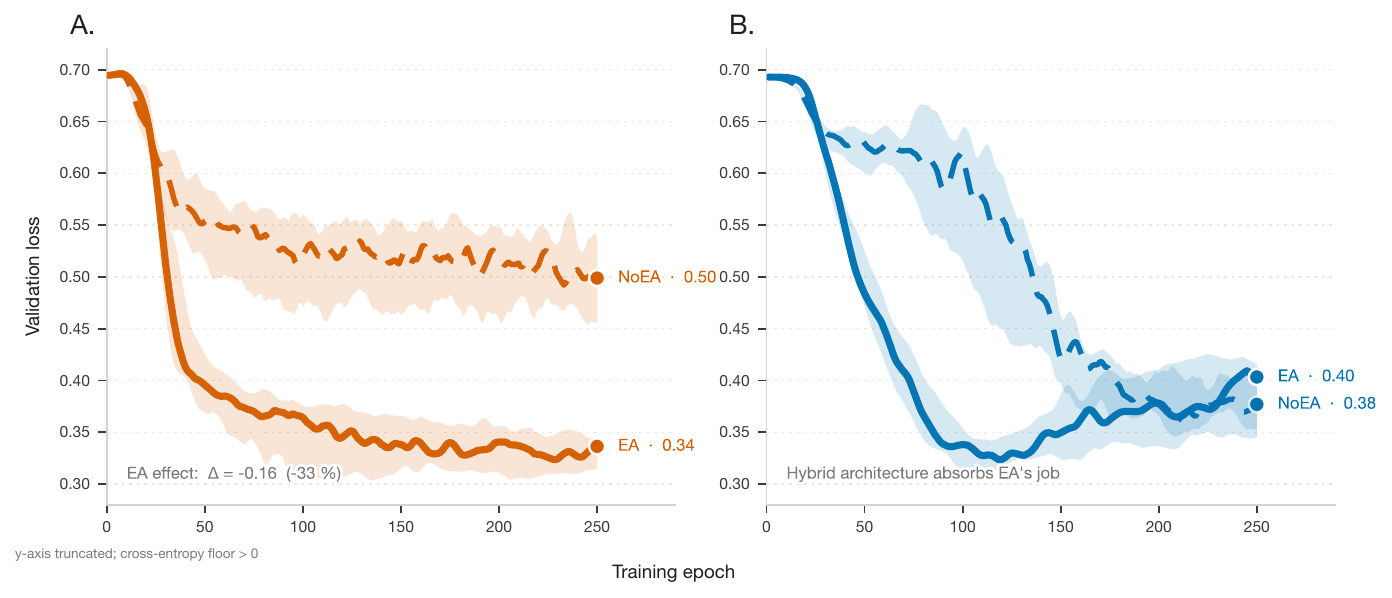}
    \caption{\textbf{The hybrid architecture takes over the role of Euclidean Alignment.}
    Validation loss per training epoch, median across 9 per-subject runs (shaded band = inter-quartile range, smoothed with a 7-epoch rolling window).
    \textbf{Left.} Shared model. Solid vermillion = EA; dashed vermillion = NoEA. EA reduces the final median loss by 33\% ($\Delta = -0.16$); without EA the loss never converges to the same level.
    \textbf{Right.} Hybrid model. Solid blue = EA; dashed blue = NoEA. The hybrid reaches essentially the same final loss with or without EA (NoEA $\approx 0.38$ versus EA $0.40$, $\Delta = +0.02$); the two curves nearly coincide, so the per-subject encoders take over EA's role. Either way the hybrid reaches the regime the shared encoder attains only with external EA.}
    \label{fig:loss_curves}
\end{figure}

Figure~\ref{fig:loss_curves} reports the validation-loss trajectories for both architectures with and without EA, smoothed by a 7-epoch rolling median over 9 per-subject runs and truncated at the largest epoch where at least half of the runs are still alive. For the shared encoder, EA is essential: applying it cuts the final median validation loss by 33\% (from 0.50 to 0.34), and removing it leaves the loss plateauing well above. For the hybrid encoder, EA has little effect on the final loss: with or without EA it converges to essentially the same level (median $0.40$ with EA versus $0.38$ without, a $+0.02$ shift), well below the shared non-aligned curve ($0.50$). The without-EA curve nearly coincides with the aligned one. That the same architecture requires EA under a shared encoder but not once each subject has its own encoder indicates that the per-subject encoders take over EA's role, learning from task supervision alone the \emph{common} representation EA otherwise imposes externally. The latent geometry below corroborates this.

\FloatBarrier

\paragraph{Latent geometry.}\label{sec:latent_geom}

\begin{figure*}[!htbp]
    \centering
    \includegraphics[width=\linewidth]{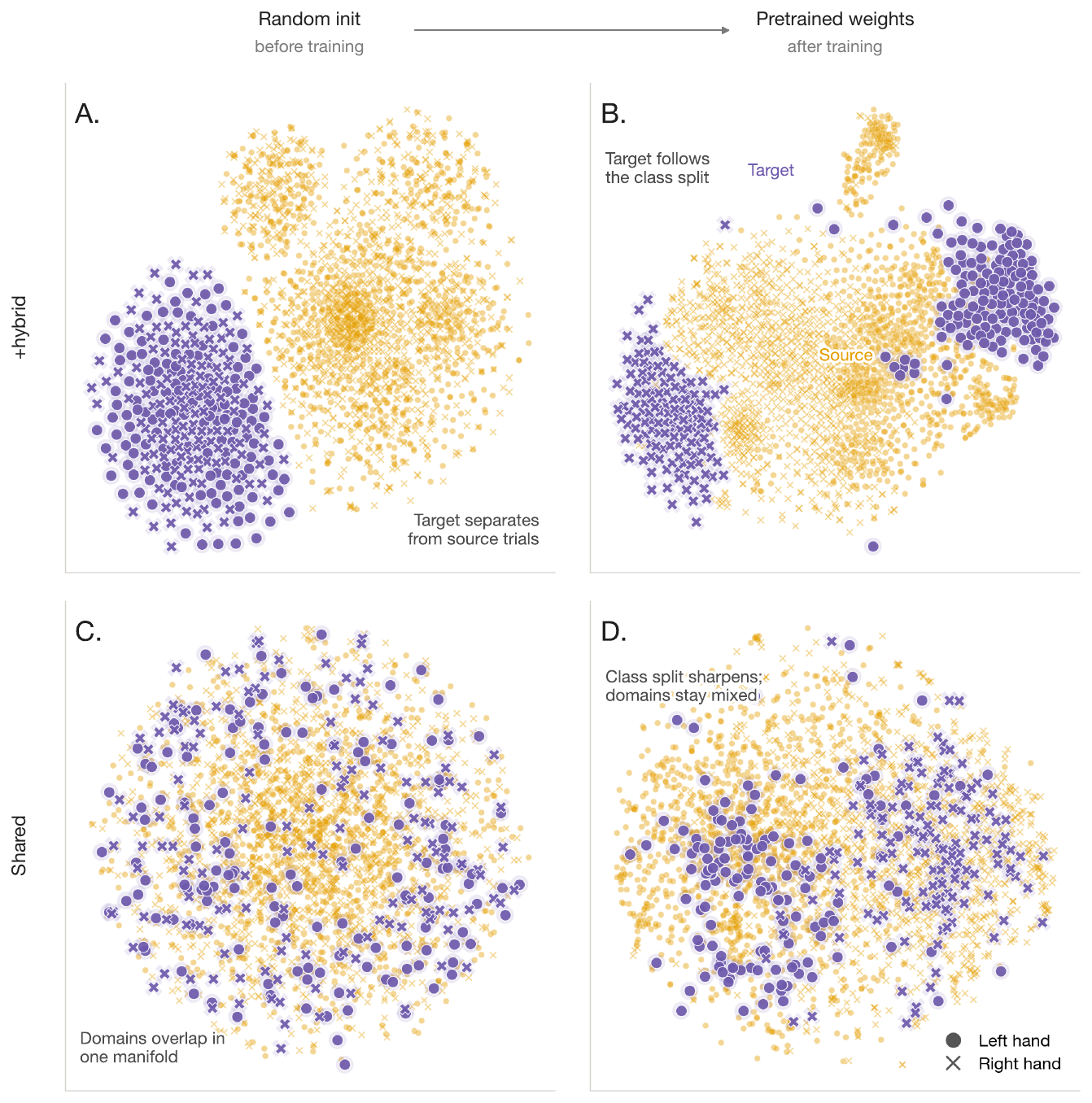}
    \caption{\textbf{Training splits the held-out subject by class.}
    Two-dimensional t-SNE of the 496-dimensional encoder features for the held-out target subject (\emph{Target}, subject~8 from \emph{BNCI2014} dataset, encoded through one source head) and the eight source subjects (\emph{Source}), under Euclidean Alignment. Rows: the hybrid encoder bank (top) and the shared encoder (bottom). Columns: random initialisation (left, the untrained control) and after training (right). Color marks domain (Source / Target), and shape marks class (circle / cross); target points are drawn larger because they are the held-out focus. Each panel is an \emph{independent} t-SNE embedding (perplexity~$=50$), so only within-panel grouping is interpretable, not distances, positions, or cluster sizes across panels. At random initialization, the hybrid encoder already places the target in its own cluster, separate from the source pool; after training the target splits by class, which cluster at the periphery of the source pool, tracking the matching source class rather than dissolving into it. The shared encoder begins with all subjects overlapping in one manifold, and training mainly sharpens the class boundary within it.}
    \label{fig:latent}
\end{figure*}

By projecting the encoder's latent features for both the aligned shared and the non-aligned hybrid architectures using t-distributed stochastic neighbor embedding (t-SNE), we can observe the impact of the training on the subjects' latent distributions. For the shared encoder, all distributions are initially projected in the same manifold, and learning makes the classes more well separated. For the hybrid encoders, however, the representations start with a strong inductive bias of separation, since all training subjects are projected differently. Figure~\ref{fig:latent} makes this contrast explicit, tracking both architectures from random initialisation to the trained model.

\begin{figure*}[!htbp]
    \centering
    \includegraphics[width=0.7\linewidth]{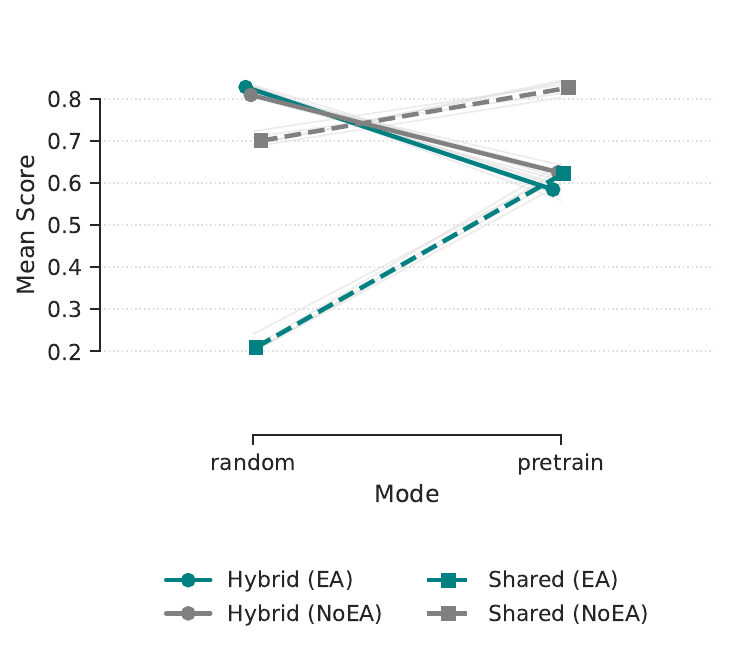}
    \caption{\textbf{Subject identity persists after training but evolves oppositely across architectures.}
    Linear subject-identity decoding accuracy from EEGNet encoder features on the BNCI2014 dataset, evaluated at random initialization (left) and after training (right), for the hybrid encoder bank and the shared encoder, with and without Euclidean Alignment (EA). Each line tracks one architecture/alignment combination from its random-init score to its trained score. The hybrid encoder loses subject identifiability with training regardless of EA, converging from $\sim0.80$ to $\sim0.60$ under both conditions. The shared encoder shows the opposite trend and is markedly EA-dependent: under EA it starts near floor ($\sim0.20$) and rises sharply after training ($\sim0.67$), while without EA it starts already substantial ($\sim0.7$) and rises more modestly ($\sim0.82$).}
    \label{fig:decoding}
\end{figure*}

\revised{Beyond this qualitative analysis, we quantify the same contrast using linear subject-identity decoding from the embeddings. Figure~\ref{fig:decoding} shows that marginal linear subject decoding remains well above chance, confirming that subject identity persists in the latent space without impeding the global classifier's decision rule. Illustrating also how this quantity evolves over training, Figure~\ref{fig:decoding} also tracks subject decoding for EEGNet from random initialization to the trained model, revealing opposite dynamics by architecture: Hybrid encoders lose subject identifiability with training irrespective of EA, with near-identical trajectories under both conditions ($\sim0.80 \to \sim0.62$). Shared encoders instead gain identifiability with training, and this gain is strongly EA-dependent: under EA, identifiability starts near floor at initialization ($\sim0.22$) and rises sharply to $\sim0.67$, whereas without EA it starts already substantial ($\sim0.70$) and rises more modestly to $\sim0.82$. These findings suggest that hybrid encoders achieve robust class-relevant subspace alignment while decreasing subject-related idiosyncrasies with training, while Shared encoders accumulate it.}


\FloatBarrier

Neither architecture receives an explicit class-separation or domain-alignment loss, so the only pressure on the latent space is the shared classifier acting on top of the encoders. Therefore, the figure highlights that, even with a strong initial bias, the hybrid model is able to reorganize the latent space using task supervision alone, making distributions closer and better separated. Although still not as aligned as the representations from the shared model, this shows that the objective is sufficient to make the subject-specific heads specialize and filter task-relevant features, while improving cross-subject alignment.

\paragraph{Synthesis.}
Two independent measurements converge on the same conclusion: validation-loss convergence (Figure~\ref{fig:loss_curves}) and inter-subject latent representation (Figure~\ref{fig:latent}) both show that the hybrid encoder \emph{internalizes a role similar to EA for the shared encoder}. The shared encoder requires EA both to converge and to bring subject distributions together; removing EA collapses both. The hybrid encoder achieves similar outcomes with or without EA, consistent with each subject's encoder performing filtering and recentering of a functional effect similar to EA. \revised{Quantitative analysis in the embeddings also demonstrate subject identity is still maintained and decodable from both Hybrid and Shared settings, although training reduces it on the Hybrid and increases for the Shared.
We do not claim that the learned encoders reproduce EA's recentering: their mechanism differs, and the convergence and latent-geometry effects above need not act at the covariance level on which EA operates.}
\revised{Regarding performance comparison with the baseline shared, however, that residual accuracy effect is small, and the substitution between the encoder bank and Euclidean Alignment is the robust finding rather than accuracy gains.}

\subsection{Subject-level heterogeneity and its implications}
\label{sec:subject_profile}

The summary measurements so far treat subjects as exchangeable units. Subject-level performance, however, is far from uniform, and the variability has more than one mode. Some subjects are consistently hard for every method; others swing widely depending on which alignment and architecture they encounter. Disentangling these two regimes matters because they call for different mitigations: a consistently hard subject suggests irreducible signal limitations, while a method-noisy subject suggests the model bank still has room to grow.

\begin{figure}[!htbp]
    \centering
    \includegraphics[width=\linewidth]{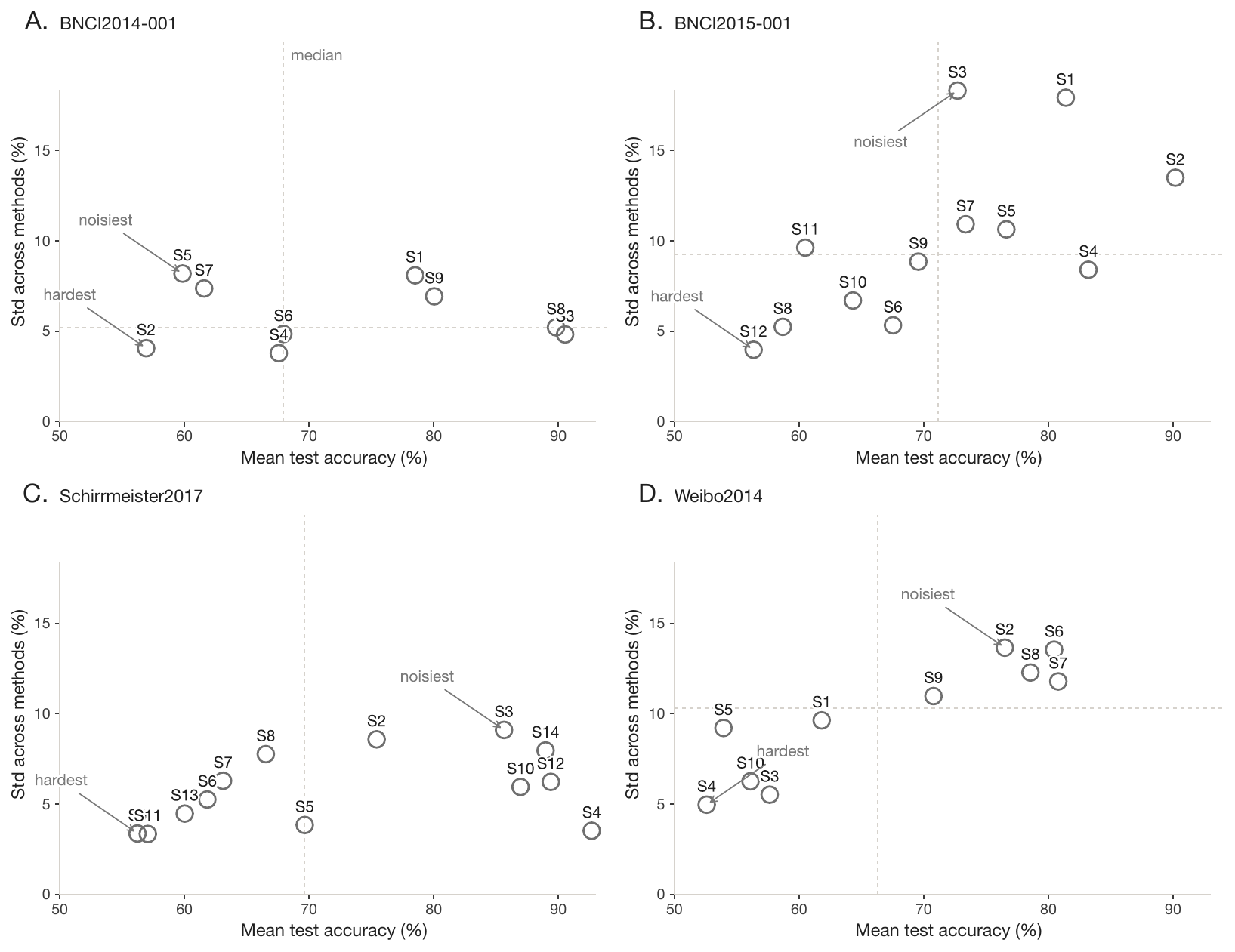}
    \caption{\textbf{Subjects split into a consistently-hard mode and a method-noisy mode, with the mix changing per dataset.}
    Each point is one subject; $x$ = mean test accuracy across all 18 method $\times$ alignment $\times$ head-selection combinations; $y$ = standard deviation of that same population.
    Dashed grey lines mark per-panel medians; the resulting quadrants read as: lower-left = consistently hard, lower-right = consistently easy, upper-half = method-noisy.
    The callouts label the lowest-mean and highest-std subjects per dataset.
    Subject behaviour is heterogeneous and at least two-dimensional: method-noisy subjects sit anywhere along $x$, including near-ceiling accuracy.}
    \label{fig:subject_profile}
\end{figure}

Figure~\ref{fig:subject_profile} plots each subject as a single point with $x =$ mean test accuracy and $y =$ standard deviation across the 18 method $\times$ alignment $\times$ head-selection combinations available per dataset (the standard deviation is our proxy for the per-subject prediction entropy the underlying experiments do not record directly). The two callouts in each panel name the most extreme subject along each axis. BNCI2014-001's subject 2 ($56.9 \pm 4.1$) is consistently hard; BNCI2015-001's subject 3 ($72.7 \pm 18.3$) is the most method-noisy; Weibo2014's subject 4 is consistently hard ($52.6 \pm 5.0$) \emph{and} its subject 2 ($76.5 \pm 13.7$) the most method-noisy. Schirrmeister2017's subjects are comparatively uniform, its noisiest reaching only $\sigma \approx 9$.

\FloatBarrier

The two regimes motivate different next steps to test. For consistently-hard subjects, none of the architectural variants we evaluated closed the gap, suggesting that improvements likely require either subject-specific calibration data or signal-quality interventions outside the current pipeline. For method-noisy subjects, the per-subject spread across configurations implies that some configuration is closer to the achievable upper bound; whether a method-ensembling or per-subject head-selection policy can recover that upper bound at test time without oracle access is the question those subjects pose for future work.

\section{Discussion and limitations}

\revised{Taken together, our results support a single thesis: per-subject encoders and Euclidean Alignment are \emph{substitutable} mechanisms for cross-subject invariance. Each recovers most of the other's benefit, so once one is in place the other adds little, and under a matched-adaptation comparison the encoder bank does not improve accuracy over a fine-tuned shared baseline. The contribution is therefore not an accuracy gain but a characterisation of this substitution.}
The experiments show that cross-subject EEG decoding is limited less by classifier capacity than by how each subject is represented before the shared decision rule.
EA improves the shared encoder because it removes a dominant covariance-level shift before learning begins, which otherwise would cause negative transfer and strongly limit the capacity of the classifier to find a class-discriminant decision boundary.
The hybrid architecture targets the same bottleneck inside the model: each subject-specific encoder learns a transformation that makes its trials more compatible with a common classifier, as shown by the loss and latent representation analyses. The class distinctiveness and relative distances show additional benefits of applying EA on top of the hybrid model, such as bigger class separation on training subjects and easier target adaptation.

This does not mean that the learned encoders completely reproduce the EA pointwise. Rather, they appear to be working under a similar mechanism, induced by the pressure of the architecture and the objective: bringing representations closer while keeping classes well separated. The class-distinctiveness and head-specificity analyses further show that the gains are not explained by adding generic capacity; instead, they depend on routing a subject through an encoder whose learned manifold matches that subject.

Comparing the evolution of the latent representations shows that the sole pressure of this learning mechanism wins out over the inductive bias of subject separation, making the representations closer in space, although still not as unified as on the aligned shared baseline, and the classes better separated. 

Although the model architecture and supervised objective together can be driven to learn subject-specific preprocessing transformations, this study has limitations in both analysis and deployment.

First, the individual encoders learn subject-specific characteristics, obtaining individualized weight representations of the training distributions. However, this specialization limits its applicability to test-time inference on a new subject. For unseen subjects, the model must select or adapt an initialization from limited calibration data, and the subject-profile analysis shows that this problem is heterogeneous: some subjects remain consistently hard, while others are sensitive to method choice. Future work should therefore focus on non-oracle head selection, calibration-efficient adaptation, and diagnostics that identify whether a target subject needs more data or a better-matched initialization. 

Although this work focused on cross-subject decoding, the same head specialization could improve temporal robustness in a cross-session setting, as studied by Wei et al.~\cite{Wei2021}. The representation visualizations show a spread and separated latent space for the training subjects, meaning that performance could largely benefit from that. Later work should investigate the benefits of using this type of architecture on additional data from training subjects.

Related ideas appear in recent non-invasive \emph{speech} decoding, where a shared encoder is equipped with an early, per-subject \emph{linear} layer, a channel-wise transform that maps each subject into a common latent space while the rest of the network stays shared~\cite{defossez2023decoding, dascoli2024decoding, levy2025braintotext}. Ablating this layer markedly degrades decoding~\cite{defossez2023decoding}, confirming that subject-dependent effects must be modelled explicitly. Our design takes the \emph{opposite} parameterisation: rather than a single linear block on top of a shared encoder, we learn a full per-subject \emph{encoder} feeding one shared classifier, and the loss and latent-geometry analyses (Figures~\ref{fig:loss_curves} and~\ref{fig:latent}) show that these encoders reach the same latent organisation with or without EA, taking over its alignment role through learned feature transformations rather than the explicit covariance recentring EA applies. This is an alignment role those subject layers assume but never test. This expressiveness comes at a cost: full per-subject encoders grow far faster than a per-subject matrix as the number of training subjects increases, which is not scalable to more diverse datasets and motivates parameter-efficient alternatives such as subject-conditioned low-rank adapters~\cite{klein2025mitigating}\revised{, which we evaluate under our held-out protocol and find to match the full encoder across ranks and backbones (Section~\ref{sec:results}, Table~\ref{tab:lora_rank})}. It also leaves the same open problem these speech models do not address: none evaluates a held-out subject, which is precisely the deployment bottleneck that our head-selection and calibration analysis targets.

Two limitations bound these conclusions. First, deploying the model to a new subject still depends on selecting or adapting a head from limited calibration data, a problem the subject-profile analysis shows is heterogeneous across subjects. Second, dedicating one encoder per subject does not scale to large or diverse subject pools, which motivates the parameter-efficient alternatives discussed above.

\section*{Data availability}

The datasets analysed in this study are publicly available and were accessed through \textsc{MOABB}~\cite{Aristimunha_Mother_of_all_2023}: BNCI2014-001 [\href{https://www.frontiersin.org/journals/neuroscience/articles/10.3389/fnins.2012.00055/full}{https://www.frontiersin.org/journals/neuroscience/articles/10.3389/fnins.2012.00055/full}]~\cite{BNCI2014001}, BNCI2015-001 [\href{https://ieeexplore.ieee.org/document/6177271/}{https://ieeexplore.ieee.org/document/6177271/}]~\cite{BNCI2015}, High-Gamma [\href{ https://onlinelibrary.wiley.com/doi/10.1002/hbm.23730}{ https://onlinelibrary.wiley.com/doi/10.1002/hbm.23730}]~\cite{Schirrmeister2017}, and Weibo2014 [\href{https://journals.plos.org/plosone/article?id=10.1371/journal.pone.0114853}{https://journals.plos.org/plosone/article?id=10.1371/journal.pone.0114853}~\cite{Yi2014EEGMotorImagery}].

\bibliography{sample}

\section*{Code availability}
The code used to train the models and to produce the analyses and figures is publicly available at \fillme{\url{https://github.com/brunaafl/LatentHybrid}}. 

\section*{Competing interests}

Bruno Aristimunha has been associated with Yneuro since November 2025. The other authors declare no competing interests.

\section*{Funding}
BJL, GS, and RYC thank the São Paulo Research Foundation (FAPESP) for the financial support (grants 22/08920-0, 23/06407-7, and 21/12645-2). 

\section*{Author contributions}
B. J. Lopes and B. Aristimunha wrote the main manuscript and prepared the figures.  B. J. Lopes and G. Schwartz developed and implemented the methodological pipeline. R. Y. de Camargo, B. Aristimunha and S. Chevallier contributed for the conception and design and R. Y. de Camargo and S. Chevallier provided critical revision of the methodology and interpretation. B Aristimunha and R. Y. de Camargo were responsible for the supervision. All authors reviewed and approved the final manuscript.

\end{document}